\documentclass[review]{elsarticle}

\usepackage{geometry}
\usepackage{graphicx}
\usepackage{diagbox}
\usepackage{subfigure}
\usepackage{multirow}
\usepackage{multicol}
\usepackage{amsmath,amsfonts,amssymb}
\usepackage{url}
\usepackage{csquotes}
\usepackage[ruled,vlined]{algorithm2e}
\usepackage[T1]{fontenc}
\usepackage{gensymb}
\usepackage[dvipsnames]{xcolor}

\newcommand{\PreserveBackslash}[1]{\let\temp=\\#1\let\\=\temp}
\newcolumntype{C}[1]{>{\PreserveBackslash\centering}p{#1}}

\begin{document}

\begin{frontmatter}

\title{Animal Behavior Classification via Deep Learning on Embedded Systems}

\author[data61]{Reza~Arablouei\corref{cor1}}
\author[af]{Liang~Wang}%[orcid=0000-0003-3529-1680]
\author[data61]{Lachlan~Currie}
\author[data61]{Jodan~Yates}
\author[dpi]{Flavio~A.~P.~Alvarenga}
\author[af]{Greg~J.~Bishop-Hurley}%[orcid=0000-0002-7671-8292]
\cortext[cor1]{Corresponding author}
\address[data61]{Data61, CSIRO, Pullenvale QLD 4069, Australia}
\address[af]{Agriculture and Food, CSIRO, St Lucia QLD 4067, Australia}
\address[dpi]{NSW Department of Primary Industries, Armidale NSW 2350, Australia}

\begin{abstract}

We develop an end-to-end deep-neural-network-based algorithm for classifying animal behavior using accelerometry data on the embedded system of an artificial intelligence of things (AIoT) device installed in a wearable collar tag. The proposed algorithm jointly performs feature extraction and classification utilizing a set of infinite-impulse-response (IIR) and finite-impulse-response (FIR) filters together with a multilayer perceptron. The utilized IIR and FIR filters can be viewed as specific types of recurrent and convolutional neural network layers, respectively. We evaluate the performance of the proposed algorithm via two real-world datasets collected from total eighteen grazing beef cattle using collar tags. The results show that the proposed algorithm offers good intra- and inter-dataset classification accuracy and outperforms its closest contenders including two state-of-the-art convolutional-neural-network-based time-series classification algorithms, which are significantly more complex. We implement the proposed algorithm on the embedded system of the utilized collar tags' AIoT device to perform in-situ classification of animal behavior. We achieve real-time in-situ behavior inference from accelerometry data without imposing any strain on the available computational, memory, or energy resources of the embedded system.

\end{abstract}


\end{frontmatter}

\section{Introduction}

The term behavior is commonly used by animal scientists to describe what an animal does during its daily life. It defines the internally coordinated responses of living organisms to internal or external stimuli~\cite{beh}. Animal behavior, when considered over appropriate periods of time, is an important indicator of health, welfare, and productivity, particularly for livestock. It can also provide valuable information about animals' environment, social interactions, and herd dynamics. % excluding those associated with developmental changes

Manual observation and recording of animal behavior is laborious and in some cases impractical. In addition, employing machine learning algorithms based on computer vision or sound recognition to automate animal behavior classification is challenging. That is because, apart from the technical challenges involved, the limited coverage range of typical fixed vision or sound sensors makes them unsuitable for monitoring large numbers of animals spread over large areas. Therefore, classifying animal behavior on wearable devices, such as small and light smart tags, using inertial measurement data is highly desirable. Micro-electro-mechanical accelerometers are compact and low-power motion sensors that can measure acceleration on three orthogonal spatial axes by sensing minute variations in the capacitance between a fixed electrode and a proof mass due to any force applied to the sensor. There is a vast body of literature around using accelerometry data to classify various animal behaviors, e.g., see~\cite{ca1}--\cite{em4aa5} and the references therein. However, there is relatively little work reported on performing the behavior classification on the embedded system of the device containing the accelerometer, e.g.,~\cite{ca10}--\cite{em4aa5}, as most of the processing is conventionally done after collecting the data.

Collecting and storing raw accelerometry data for post-hoc processing is inefficient and unscalable. Transferring the raw data via wireless communication is similarly disadvantageous. Therefore, it is important to realize the classification of animal behavior in-situ and in real-time on the embedded system of the wearable device that collect the data. Doing so, only the inferred behavior classes need be stored or communicated.

In this paper, we develop a deep-neural-network-based supervised machine-learning algorithm to classify animal behavior using accelerometry data on the embedded system of a custom-built artificial intelligence of things (AIoT) device that can be worn by cattle and similar livestock as a collar tag. The proposed algorithm can effectively be used for behavior inference on the embedded system of the AIoT device without straining its computational, memory, or energy resources.

Most existing animal behavior classification algorithms are conventional feature-engineering-based approaches that involve separate feature extraction and classification processes. It is common to take various time- and frequency-domain statistics and measures as features, for example, mean, standard deviation, skewness, kurtosis, maximum value, minimum value, autocorrelation, median, median absolute deviation, dominant frequency, and entropy. Some other rather ad-hoc values, such as the so-called overall dynamic body acceleration and vectorial dynamic body acceleration~\cite{odba}, have also been used as features. The main drawback of such approaches is that the features are pre-defined regardless of the classifier used and need be carefully engineered and hand-picked possibly through a suitable feature selection method. The engineered features are also often limited in flexibility and utility.

Our new animal behavior classification algorithm is composed of two main processes that can be viewed as performing feature calculation and classification. However, it enjoys end-to-end learning since the feature calculation process contains learnable parameters that are trained jointly with the parameters of the classifier. Therefore, the algorithm does not rely on any hand-engineering of the features as it learns them directly from the data.

The proposed algorithm extracts meaningful and computationally efficient features that facilitate classification of animal behavior in-situ and in real-time on the embedded system of the collar tag's AIoT device. To this end, in the proposed algorithm, we use a set of first-order infinite-impulse-response (IIR) Butterworth high-pass filters and a set of nonlinear filters composed of two linear finite-impulse-response (FIR) filters joined by tangent hyperbolic nonlinear activation. To enable end-to-end learning of the deep neural network model defining the proposed algorithm, we make the parameters of the utilized IIR and FIR filters learnable. We design the proposed algorithm with the aim of performing inference using the learned models on the embedded system of the collar tag's AIoT device. Therefore, we take into account the computational, memory, and energy constraints of the embedded system.

We carry out model training on a suitable computing device using a deep-learning library where we implement the IIR and FIR filters employed for feature calculation as specific recurrent and convolutional neural networks, respectively. We then deploy the learned model on the embedded system using a library provided by the microcontroller manufacturer.

We evaluate the performance of the proposed algorithm using two real-world datasets containing accelerometry data collected from grazing beef cattle and annotated manually. The proposed algorithm exhibits excellent intra- and inter-dataset classification accuracy and outperforms two state-of-the-art convolutional-neural-network(CNN)-based algorithms recently proposed for end-to-end classification of time-series with a considerably smaller time and memory complexity.

We also provide some insights into how the proposed algorithm works by analyzing the statistical and spectral properties of the accelerometry data and the characteristics of the extracted features.

\section{Data}

In this section, we describe the procedures and tools used to generate two datasets that we consider in this work, i.e., data collection experiments, utilized hardware, annotation process, and data segmentation.

\subsection{Experiments}

We have obtained our datasets from grazing beef cattle of Angus breed during two data collection experiments ran in August 2018 and March 2020. The first experiment took place in August 2018 for 28 days at the Commonwealth Scientific and Industrial Research Organisation (CSIRO) FD McMaster Laboratory Pasture Intake Facility~\cite{ts_armidale}, Chiswick NSW, Australia (30\degree36'28.17"S, 151\degree32'39.12"E). The accelerometry data was collected from ten steers wearing collar tags called eGrazor\footnote{\url{https://www.csiro.au/en/research/animals/livestock/egrazor-measuring-cattle-pasture-intake}}. The steers were 23 to 35 months of age and weighed 530 to 816 kg. We refer to the associated dataset as Arm18. Another experiment was conducted in March 2020 for eight days at the same facility while the accelerometry data was recorded from eight heifers wearing the eGrazor collar tags. The heifers were 19 months old and weighed 283 to 354 kg. We refer to the associated dataset as Arm20. In Table~\ref{tab:weather}, we provide summary statistics of the weather conditions for the periods that the experiments took place at Armidale, NSW, Australia\footnote{\url{http://www.weatherarmidale.com/}}.

In both experiments, the cattle wore the collar tags uninterruptedly. Therefore, the eGrazor collar tags logged the accelerometry data continuously for the entire duration of the experiments. At the conclusion of each experiment, we retrieved the SD flash memory cards, which stored the logged data, from the tags. There was no concern around the storage capacity as, with a sampling rate of 50 readings per second, a 32GB memory card can accommodate the IMU data of about 400 days.

\newcolumntype{L}{>{\centering\arraybackslash}m{3.8cm}}
\begin{table} \footnotesize
\caption{Weather summary statistics during the experiments at Armidale NSW.\\}
\label{tab:weather}
\vspace{10pt}
\centering
\begin{tabular}{|l| L L|} % p{3.7cm} p{3.7cm}
\hline
\backslashbox{condition\qquad}{time} & August 2018 & March 2020\\
\hline
average maximum temperature   & 15.7{\degree}C & 22.3{\degree}C\\
average minimum temperature   & -1.9{\degree}C & 11.5{\degree}C\\
highest maximum temperature   & 20.6{\degree}C & 28.6{\degree}C\\
lowest minimum temperature    & -7.6{\degree}C & 3.6{\degree}C\\
\hline
maximum relative humidity     & 89\%           & 91\%\\
minimum relative humidity     & 48\%           & 55\%\\
\hline
average pressure              & 1018 mbar      & 1017 mbar\\
\hline
average daily rainfall        & 0.9 mm         & 1.9 mm\\
average daily pan evaporation & 2.3 mm         & 2.9 mm\\
\hline
overall                       & dry, sunny, and frosty with cool days and cold nights 
                              & warm, cloudy, and damp with mostly light falls of rain\\
\hline
\end{tabular}
\end{table}

Fig.~\ref{pad} shows the paddock and the cattle used for the experiment that produced the Arm20 dataset. Figs.~\ref{axes} shows cattle wearing the eGrazor collar tags. The experiments were approved by the CSIRO FD McMaster Laboratory Chiswick Animal Ethics Committee with the animal research authority numbers 17/20 and 19/18.

\subsection{eGrazor}

\begin{figure*}
    \centering
    \subfigure[Cattle on paddock during the experiment resulting in the Arm20 dataset.]{\includegraphics[height=5.8cm]{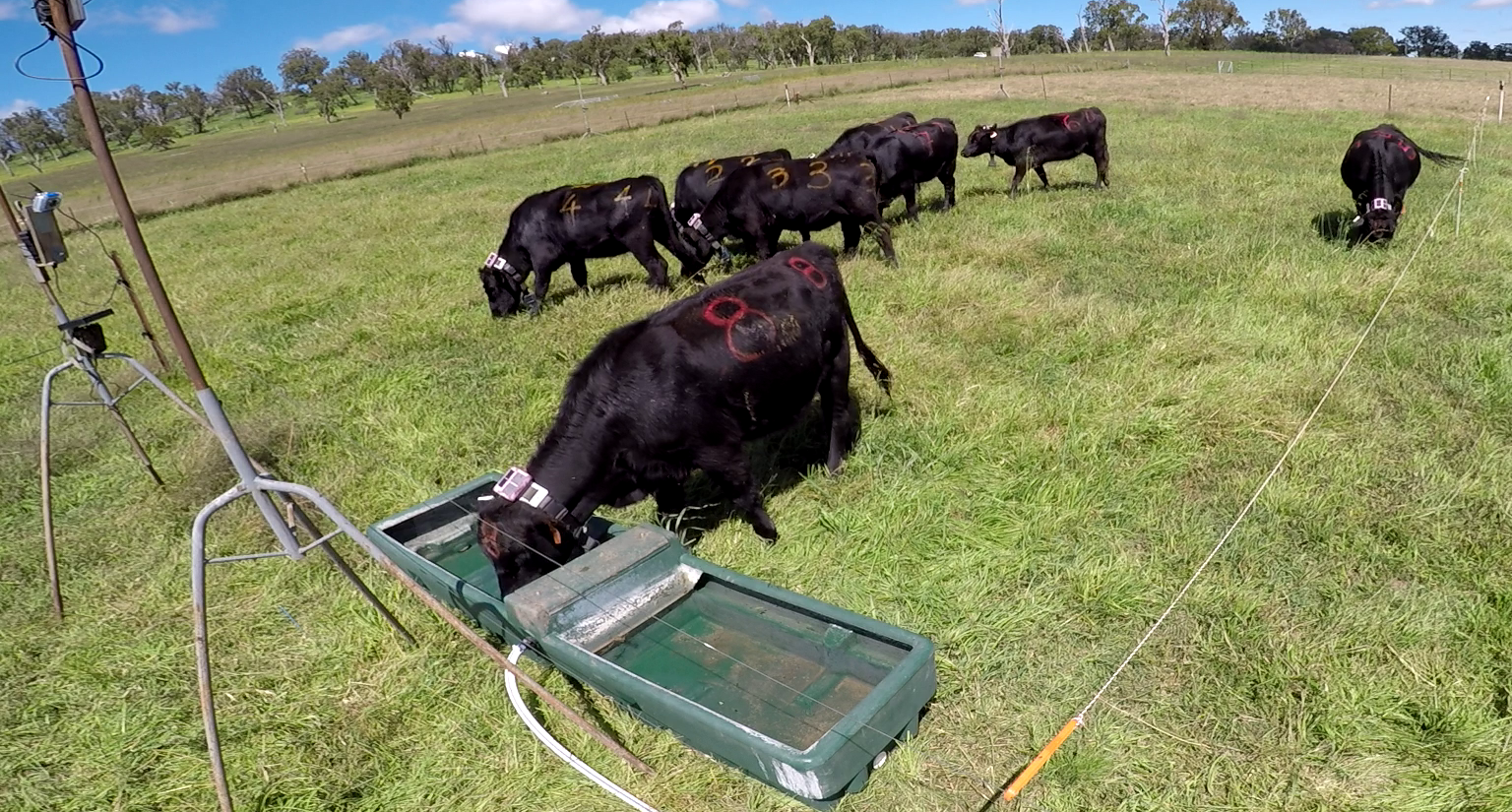}\label{pad}}\vspace{.4cm}
    \subfigure[Cattle wearing eGrazor collar tags and the three spatial axes of the utilized triaxial accelerometer.]{\includegraphics[height=5.8cm]{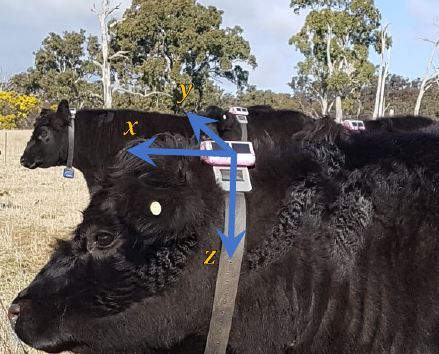}\label{axes}}\vspace{.4cm}
    \subfigure[eGrazor collar tag including Loci, battery pack, and solar panels.]{\includegraphics[height=5.05cm]{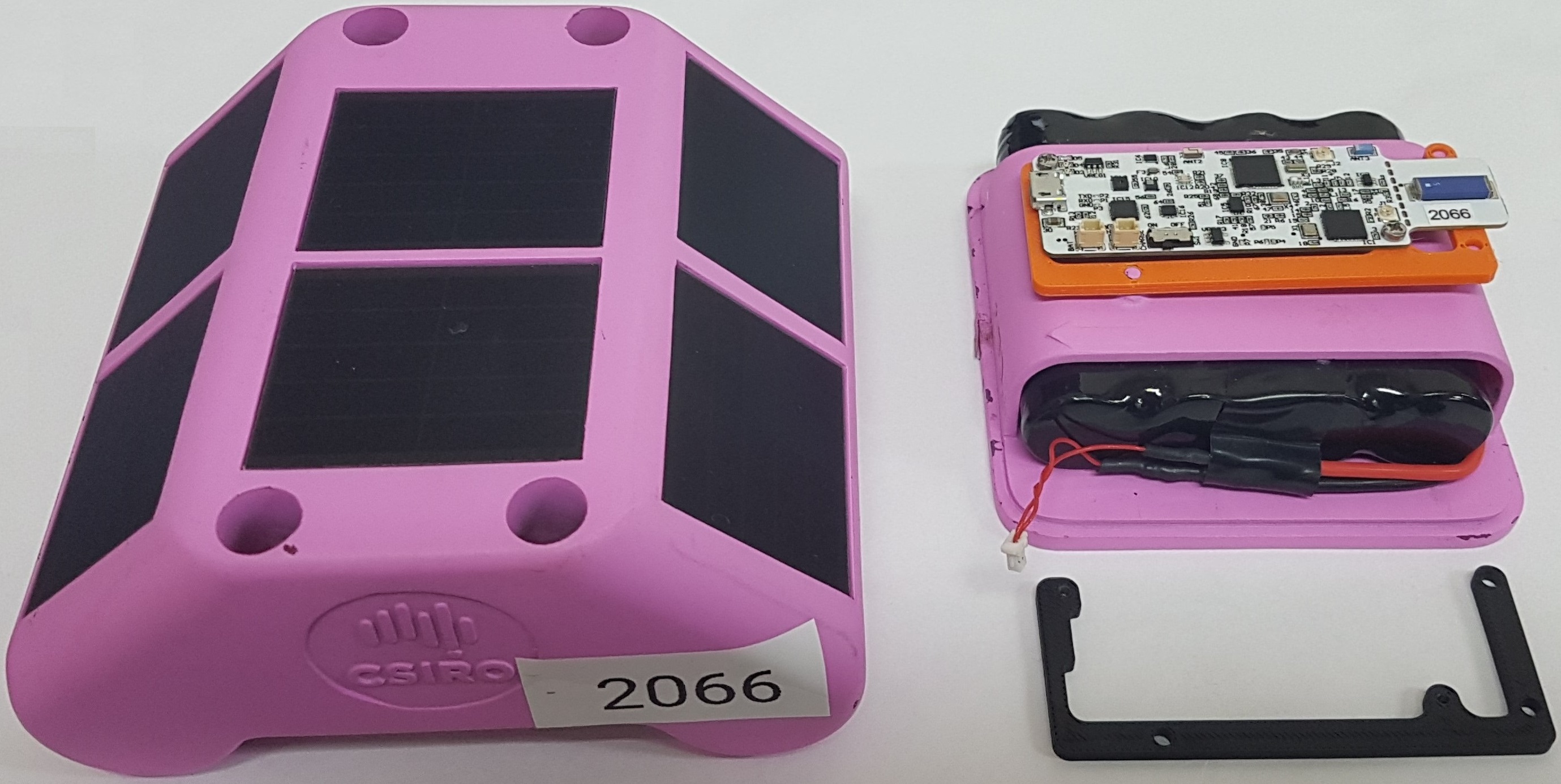}\label{tag}}\vspace{.4cm}
\end{figure*}

During the experiments, we fitted the cattle with our eGrazor collar tags that are purpose-built to capture, log, and process various sensor data including inertial measurement, temperature, pressure, and geo-location using the global navigation satellite system (GNSS). The tag, shown in Fig. \ref{tag}, houses an artificial intelligence of things (AIoT) device called Loci, a battery pack, and six photovoltaic modules for harvesting solar energy. We place the tag on top of the animal's neck and secure it with a strap and a counterweight.

\begin{figure*}
    \centering
    \subfigure[Loci, the AIoT device used for data collection and in-situ behavior classification.]{\includegraphics[height=5.05cm]{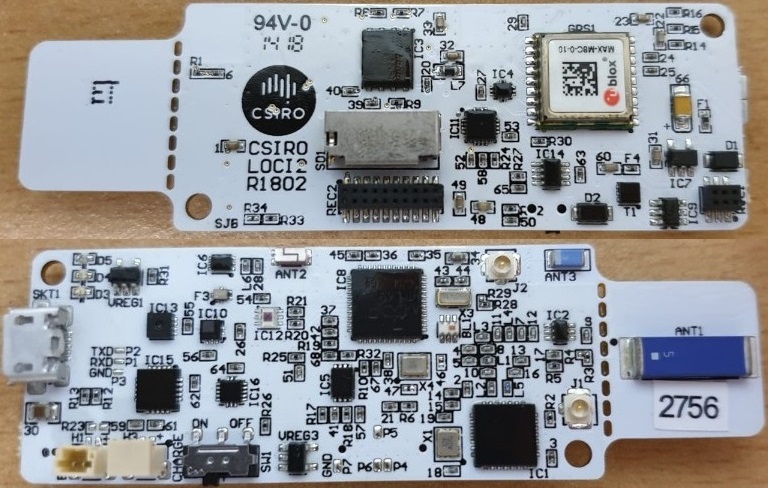}\label{loci}}
    \caption{\small{The experiment paddock containing cattle wearing eGrazor collar tags and images of eGrazor and its AIoT device, Loci, used for collecting accelerometry data corresponding to cattle behavior.}}
    \label{tag-cow}
\end{figure*}

Loci, shown in Fig.~\ref{loci}, contains a wealth of sensing and communication capabilities. It has a Texas Instruments CC2650F128 system-on-chip that consists of an Arm Cortex-M3 CPU running at $48$MHz with $28$KB of random access memory (RAM), $128$KB of read-only memory (ROM), and a 802.15.4 radio module. Loci also has an MPU9250 $9$-axis micro-electro-mechanical (MEMS) inertial measurement unit (IMU) including a tri-axial accelerometer sensor that measures acceleration in three orthogonal spatial directions (axes) as shown in Fig. \ref{axes}. The $x$ axis corresponds to the antero-posterior (forward/backward) direction, the $y$ axis to the medio-lateral (horizontal/sideways) direction, and the $z$ axis to the dorso-ventral (upward/downward) direction. The IMU chip outputs the tri-axial accelerometer readings as $12$-bit signed integers at a rate set to $50$ samples per second. The raw accelerometer readings can be processed by the on-board microcontroller or recorded on an external flash memory card.

The power to Loci is supplied by the $3.6$V, $13.4$Ah Lithium-ion battery pack that is recharged via six solar panels installed on the exterior of the tag case. Loci draws a maximum current of $30$mA even when the CPU and all other main components including the GNSS receiver runs continuously. Therefore, the tag can operate normally for at least $18$ days using the battery pack's full capacity with no recharge. In practice, the battery pack is recharged for several hours almost every day with solar panels providing up to $300$mA in total.


\subsection{Annotations}

We have annotated parts of the collected accelerometry data by monitoring the behavior of the cattle on the field during the 2018 experiment and viewing the video recordings of the 2020 experiment. We use the annotations to create our labeled datasets called Arm18 and Arm20 corresponding to the respective data collection experiments as detailed above. We consider six mutually-exclusive behaviors of grazing, walking, ruminating, resting, drinking, and other in the Arm18 dataset. We consider the same behaviors in the Arm20 dataset except for combining the ruminating and resting behaviors to a single behavior class referred to as ruminating/resting. We combine the ruminating and resting behaviors to a single behavior class in the Arm20 dataset as it is hard to clearly distinguish these behaviors in the recorded videos. The \emph{other} behavior class is the collection of all behaviors other than the considered ones, i.e., grazing, walking, ruminating, resting, and drinking.

We consider the above cattle behaviors as they are the most important behaviors from the perspective of evaluating and monitoring productivity, feed efficiency, energetic dynamics, health, and welfare of grazing cattle. For example, the knowledge of the times and durations of a cattle’s grazing is crucial for determining its herbage dry matter intake from the pasture [26]. knowing when and for how long a cattle ruminates or rests can also help understand the health and well-being state of the animal~\cite{rumin}. Monitoring the walking behavior can be useful for measuring the animal's energy expenditure while identifying the drinking behavior is essential to ascertain the animal's access to water and hence compliance with associated regulations. It is also important to note that grazing cattle, particularly beef cattle, spend vast majority of their lives performing the considered behaviors.

We have produced our annotations partially via observing the animals during the trials and partially via reviewing the recorded videos. Annotating animal behavior is generally arduous and challenging. Particularly, it is not uncommon to overlook some instances of rare behaviors such as drinking, even for a domain expert, as they happen occasionally and in short durations. Differentiating some behaviors such as ruminating and resting can also be difficult.

\subsection{Datasets}

We create the labeled Arm18 and Arm20 datasets by dividing the relevant annotated accelerometry data into non-overlapping segments each containing $256$ consecutive triaxial readings, which are unique to the segment. The segment size of $256$ readings corresponds to about $5.12$s. Table~\ref{tab:dataset} shows the number of segments (datapoints) for each behavior class in each dataset.

To determine the optimal segment size, we experimented with various values. The results show that the segment size of 256 accelerometer readings (5.12 seconds) leads to a good balance between different competing aspects of performance, i.e., classification accuracy and time/space complexity. Larger segment sizes correspond to finer resolution in the frequency domain that may help better recognize subtle differences between the classes through IIR/FIR filtering. In addition, as the accelerometer readings are considerably noisy, calculating statistical features aggregated over longer segments can help filter out the uncertainty induced by noise more effectively. However, larger segment sizes result in fewer datapoints being available for training as well as higher computational and memory complexity of performing inference on each datapoint.

\begin{table} \footnotesize
\caption{The number of labeled $256$-sample segments (datapoints) in the considered datasets for each behavior class.\\} 
\label{tab:dataset}
\centering
\begin{tabular}{|l| c c | c|}
\hline
\backslashbox{behavior}{dataset} & Arm18 & Arm20 & total\\
\hline
grazing    & 6588 & 6156 & 12744\\
walking    & 65   & 910 & 975\\
ruminating & 2502 & \multirow{2}{*}{4080} & \multirow{2}{*}{9708}\\
resting    & 3126 & & \\
drinking   & 104  & 594 & 698\\
other      & 178  & 222 & 400\\
\hline
total      & 12563 & 11962 & 24525\\
\hline
\end{tabular}
\end{table}

\section{Algorithm}

We take an end-to-end learning approach in developing our animal behavior classification algorithm. The conventional feature-engineering-based approaches involve separate feature engineering and classification processes. However, to achieve end-to-end learning, we propose an algorithm that calculates relevant features and performs classification in conjunction. The algorithm uses trainable parameters for both feature calculation and classification, which can be optimized jointly during training.

\begin{figure*}
    %\centering
    \hspace{-2.1cm}\includegraphics[scale=.71]{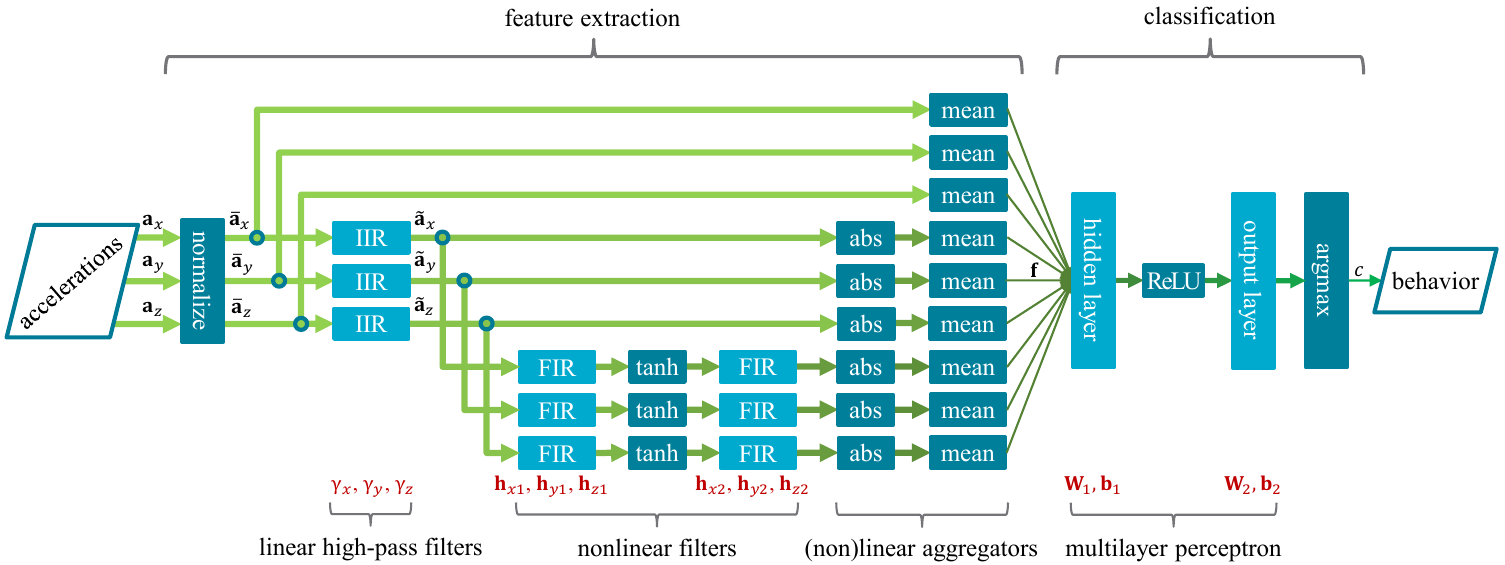}
    \caption{\small{The architecture of the model underlying the proposed animal behavior classification algorithm when performing inference. During training, the $\mathrm{argmax}$ operator is replaced with the $\mathrm{softmax}$ operator.}}
    \label{mdl}
\end{figure*}

Since we aim to realize animal behavior inference on the embedded system of Loci, we take into consideration its resource limitations in designing the underlying model of our animal behavior classification algorithm that maps triaxial accelerometry data to animal behavior classes. In Fig.~\ref{mdl}, we sketch the architecture of our proposed end-to-end animal behavior classification model. The input to the model consists of $256$ contiguous triaxial accelerometer readings and the output is the predicted animal behavior class, when performing inference. During training, the $\mathrm{argmax}$ operator in Fig.~\ref{mdl} is replaced with the $\mathrm{softmax}$ operator, whose output is used to calculate the associated cross-entropy loss.

The proposed model has two major parts, namely feature calculation and behavior classification. The main components of the feature calculation part are a set of linear high-pass IIR filters, a set of nonlinear filters each composed of two FIR filters and an element-wise hyperbolic tangent ($\mathrm{tanh}$) activation function, and corresponding mean and mean-absolute aggregation functions, which we will elaborate on in the following. The behavior classification part is made of a multilayer perceptron (MLP).

\subsection{Normalization}

We stack the triaxial accelerometer readings into three $N$-dimensional vectors, denoted by $\mathbf{a}_{x}$, $\mathbf{a}_{y}$, and $\mathbf{a}_{z}$. Recall that, in this work, we set $N=256$. Each vector contains the accelerometer readings pertaining to one spatial axis, i.e., $x$, $y$, or $z$, as signified by the associated index.

During training, we calculate the mean and standard-deviation of the accelerometer readings for each axis using the entire training data. We then normalize the accelerometer readings of each axis by subtracting the corresponding mean from them and dividing the result by the corresponding standard-deviation during both training and inference. Therefore, we express the normalized values of the accelerometer readings as

\begin{equation*}
\bar{\mathbf{a}}_d=s_d\left(\mathbf{a}_d-m_d\mathbf{1}\right),\ d\in\{x,y,z\}
\end{equation*}
where $m_d$ and $s_d$, $d\in\{x,y,z\}$, are the means and the inverse standard-deviations, respectively.

\subsection{Calculation of features}

We average the entries of $\bar{\mathbf{a}}_d$ for each axis $d\in\{x,y,z\}$ to produce three features, i.e., the mean features, as
\begin{equation}
    f_{1d}=\frac{1}{N}\mathbf{1}^{\intercal}\bar{\mathbf{a}}_d,\ d\in\{x,y,z\}
\end{equation}
where $\mathbf{1}$ stands for a column vector of appropriate size with all entries being one. As the accelerometers sense the gravity of earth, the mean features contain information about the orientation of the collar tag or equivalently the pose of animal's head.

To eliminate the effect of gravity after calculating the mean features, we filter the normalized values of the accelerometer readings of each axis using a first-order high-pass Butterworth filter that has a single adjustable parameter $\gamma_d$, $d\in\{x,y,z\}$. These IIR filters remove the low-frequency components of the normalized accelerometer readings of each axis to the extent determined by the value of $\gamma_d$. We denote the application of the utilized IIR filters by
\begin{equation}
    \left[1,-\gamma_d\right]^{\intercal}*\tilde{\mathbf{a}}_d=\left[1,-1\right]^{\intercal}*\bar{\mathbf{a}}_d,\ d\in\{x,y,z\}
\end{equation}
where $\tilde{\mathbf{a}}_d$ is the filter output and $*$ denotes the linear convolution operation. Although this notation is somewhat unorthodox, it a meaningful time-domain representation of a first-order high-pass Butterworth filter that highlights its recurrent nature and hence infinite impulse response without relying on any frequency-domain notation. The convolution on the left hand-side represents the recurrence and is required for the equality to hold. In practice, the entries of $\tilde{\mathbf{a}}_d$, $d\in\{x,y,z\}$, are calculated through recursive operations as described in~\cite{ca10}.

We compute the second set of features by averaging the absolute values of the high-pass-filtered accelerometer readings for each axis, i.e.,
\begin{equation}
    f_{2d}=\frac{1}{N}\mathbf{1}^{\intercal}|\tilde{\mathbf{a}}_d|,\ d\in\{x,y,z\}.
\end{equation}
These features contain information about the intensity of the animal's body movements. We use the mean-absolute value as a surrogate for the standard-deviation since it is more computationally-efficient and robust to noise or outliers.

The features $f_{1d}$ and $f_{2d}$, $d\in\{x,y,z\}$, are similar to the ones used in~\cite{ca10} but are different in two major aspects. First, here, the IIR filter parameters $\gamma_d$, $d\in\{x,y,z\}$, are specific to each axis while, in~\cite{ca10}, the same parameter is used for all axes. Second, unlike in~\cite{ca10} where the parameter of the IIR filters is treated as a hyperparameter, in this work, we consider $\gamma_d$, $d\in\{x,y,z\}$, to be trainable parameters whose optimal values can be learned from the data via training.

To enhance the discriminative power of the proposed model, we extract three additional features from the high-pass-filtered accelerometer readings $\tilde{\mathbf{a}}_d$. Thus, we pass them through a set of nonlinear filters each consisting of two tandem FIR filters with an element-wise $\mathrm{tanh}$ activation function in between. We then calculate the mean-absolute of the nonlinear filter outputs as the third set of features, i.e.,
\begin{equation}
    f_{3d}=\frac{1}{N}\mathbf{1}^{\intercal}|\tanh\left(\tilde{\mathbf{a}}_d*\mathbf{h}_{1d}\right)*\mathbf{h}_{2d}|,\ d\in\{x,y,z\}
\end{equation}
where $\mathbf{h}_{1d}\in\mathbb{R}^{K_1\times 1}$ and $\mathbf{h}_{2d}\in\mathbb{R}^{K_2\times 1}$ represent the impulse responses of the utilized FIR filters for each axis $d\in\{x,y,z\}$ with lengths $K_1$ and $K_2$, respectively. We treat these impulse responses as trainable parameter vectors.

Similar to the second set of features $f_{2d}$, $d\in\{x,y,z\}$, the third set of features $f_{3d}$, $d\in\{x,y,z\}$, also contain information regarding the intensity of animal’s body movements that are sensed by the accelerometers. However, the movements whose intensity is captured through $f_{3d}$, $ d\in\{x,y,z\}$, relate to specific parts of the frequency spectrum ascertained by the FIR filter coefficients, which are learned directly from the data. Here, we consider a single set of nonlinear filters and consequently one set of the associated features. However, extending the proposed algorithm to include more nonlinear filters in parallel and thus more features is straightforward.

We stack the calculated features, i.e., $f_{i,d}$, $i\in\{1,2,3\}$ $\&$ $d\in\{x,y,z\}$, in the feature vector denote by $\mathbf{f}$.

\subsection{Classification}

We feed the feature vector $\mathbf{f}$ into to an MLP that outputs $C$ numbers each corresponding to one behavior class. The employed MLP classifier has one hidden layer that is followed by the rectified linear unit (ReLU) activation function. Therefore, the output layer produces
\begin{equation}
    \mathbf{W}_{2}\max\left(\mathbf{0},\mathbf{W}_{1}\mathbf{f}+\mathbf{b}_{1}\right)+\mathbf{b}_{2}
    \end{equation}
where $\mathbf{0}$ denotes a vector of appropriate size with all zero entries, $\mathbf{W}_1\in\mathbb{R}^{L\times F}$ and $\mathbf{b}_1\in\mathbb{R}^{L\times 1}$ are the weight matrix and the bias vector of the hidden layer, $\mathbf{W}_2\in\mathbb{R}^{C\times L}$ and $\mathbf{b}_2\in\mathbb{R}^{C\times 1}$ are the weight matrix and the bias vector of the output layer, $F$ is the number of features, $L$ is the dimension of the hidden layer output, and $C$ is the number of classes. 

During training, we use the $\mathrm{softmax}$ operator to transform the output of the MLP to the pseudo-likelihoods of the considered behavior classes, which are used to calculate the associated cross-entropy loss. When performing inference with any trained model using the proposed algorithm, we choose the behavior class that has the highest corresponding MLP output.

We summarize the procedure of performing inference using the proposed algorithm together with the involved parameters and variables in Algorithm~\ref{alg}.

\begin{algorithm}[!t]\label{alg} \footnotesize
\renewcommand{\AlCapSty}[1]{\normalfont\small{#1}\unskip}
\caption{The inference procedure using the proposed algorithm and the involved parameters and variables.}
%\linespread{1.4}\selectfont
\footnotesize
\DontPrintSemicolon

\begin{tabular}{@{}l l}

input, $\forall d\in\{x,y,z\}$:\vspace{2pt}\\
\quad $\mathbf{a}_d\in\mathbb{R}^{N\times 1}$ & vectors of accelerometer readings \vspace{4pt}\\

output:\vspace{2pt}\\
\quad $c\in\{0,\dots,C-1\}$ & predicted behavior class index \vspace{4pt}\\

parameters, $\forall d\in\{x,y,z\}$:\vspace{2pt}\\
\quad $N\in\mathbb{Z}^+$ & segment length\\
\quad $K_1,K_2\in\mathbb{Z}^+$ & FIR filter lengths\\
\quad $F\in\mathbb{Z}^+$ & number of features\\
\quad $L\in\mathbb{Z}^+$ & hidden layer dimension\\
\quad $C\in\mathbb{Z}^+$ & number of classes\\
\quad $m_d\in\mathbb{R}$  & normalization means\\
\quad $s_d\in\mathbb{R}^+$  & normal. inverse standard-deviations\\
\quad $0\le\gamma_d\in\mathbb{R}^+\le1$ & IIR filter coefficients\\
\quad $\mathbf{h}_{1d}\in\mathbb{R}^{K_1\times 1},\mathbf{h}_{2d}\in\mathbb{R}^{K_2\times 1}$ & FIR filter impulse responses\\
\quad $\mathbf{W}_{1}\in\mathbb{R}^{L\times F},\mathbf{W}_{2}\in\mathbb{R}^{C\times L}$ & MLP weights\\
\quad $\mathbf{b}_{1}\in\mathbb{R}^{L\times 1},\mathbf{b}_{2}\in\mathbb{R}^{C\times 1}$ & MLP biases \vspace{4pt}\\

variables:\vspace{2pt}\\
\quad $\mathbf{f}\in\mathbb{R}^{F\times1}$ & features \vspace{4pt}\\

\end{tabular}
\vspace{2pt}\\
inference procedure:\vspace{2pt}\\
\quad normalize, $\forall d\in\{x,y,z\}$:
\begin{equation*}
\bar{\mathbf{a}}_d=s_d\left(\mathbf{a}_d-m_d\mathbf{1}\right)
\end{equation*}\\
\quad calculate features, $\forall d\in\{x,y,z\}$:
\begin{align*}
f_{1d}&=\tfrac{1}{N}\mathbf{1}^{\intercal}\bar{\mathbf{a}}_d\\
\left[1,-\gamma_d\right]^{\intercal}*\tilde{\mathbf{a}}_d&=\left[1,-1\right]^{\intercal}*\bar{\mathbf{a}}_d\\
f_{2d}&=\tfrac{1}{N}\mathbf{1}^{\intercal}|\tilde{\mathbf{a}}_d|\\
f_{3d}&=\tfrac{1}{N}\mathbf{1}^{\intercal}|\tanh\left(\tilde{\mathbf{a}}_d*\mathbf{h}_{1d}\right)*\mathbf{h}_{2d}|
\end{align*}\\
\quad classify:
\begin{align*}
c&=\arg\max\left(\mathbf{W}_{2}\max\left(\mathbf{0},\mathbf{W}_{1}\mathbf{f}+\mathbf{b}_{1}\right)+\mathbf{b}_{2}\right)
\end{align*}

\end{algorithm}

\section{Evaluation}

We evaluate both intra-dataset and inter-dataset classification performance of the proposed algorithm using our labeled datasets and appropriate cross-validation schemes. We also tune the hyperparameters of the proposed algorithm in each scenario through cross-validation and a greedy method.

We use the Matthews correlation coefficient (MCC)~\cite{mcc} for evaluating the classification accuracy. The MCC takes into account true and false positives and negatives and is known to be a meaningful measure even when the dataset is highly imbalanced. It falls between $-1$ and $+1$ where $+1$ is perfect prediction, $0$ no better than random prediction, and $-1$ perfect inverse prediction.

We jointly optimize the feature calculation parameters (the IIR and FIR filter coefficients), i.e., $\gamma_{d}$, $\mathbf{h}_{d1}$, and $\mathbf{h}_{d2}$, $d\in\{x,y,z\}$, and the MLP classifier parameters, i.e., $\mathbf{W}_1$, $\mathbf{b}_1$, $\mathbf{W}_2$, and $\mathbf{b}_2$. To this end, we implement the proposed model and train it using the PyTorch library\footnote{\url{https://pytorch.org/
}}. We use an approach similar to the one taken in~\cite{diff_iir} to implement the IIR filters with differentiable parameters. To implement the FIR filters, we use one-dimensional convolution operations with no bias or padding and set the stride to one and the number of groups to the number of input channels, i.e., three.

\subsection{Intra-dataset Accuracy}\label{subsec:ida}

We consider three datasets for evaluating the intra-dataset classification accuracy of the proposed algorithm. They are the original six-class Arm18 and five-class Arm20 datasets plus a five-class version of the Arm18 dataset. We create the five-class Arm18 dataset by combining the ruminating and resting behavior classes of the original dataset into a single ruminating/resting behavior class. Hence, we make a version of the Arm18 dataset that has the same behavior classes as the Arm20 dataset. This facilitates performance evaluation and comparison, especially, in the next subsection where we perform inter-class performance accuracy evaluation.

To evaluate the classification accuracy of the proposed algorithm with each considered dataset, we use a leave-one-animal-out cross-validation scheme. In each cross-validation fold of this scheme, we use the data of one animal for validation and the data of the other animals for training. We aggregate the results of all folds to calculate the cross-validated results.

We compare the accuracy of the proposed algorithm with those of four other algorithms, namely, a variant of the proposed algorithm that uses a set of linear FIR filters to calculate $f_{3d}$, $d\in\{x,y,z\}$, i.e.,
\begin{equation}
    f_{3d}=\frac{1}{N}\mathbf{1}^{\intercal}|\tilde{\mathbf{a}}_d*\mathbf{h}_{d}|,\ d\in\{x,y,z\},
\end{equation}
the MLP-based algorithm of~\cite{ca10}, and two CNN-based time-series classification algorithms proposed in~\cite{fcn} and called the fully convolutional network (FCN) and the residual network (ResNet). It is shown in~\cite{dl4tsc} that FCN and ResNet are among the most accurate existing time-series classification algorithms, specifically those based on deep learning.

We utilize the Adam algorithm~\cite{adam} to optimize the cross-entropy loss associated with the multiclass classification problem. We tune the model and training hyperparameters of the proposed algorithm for each dataset in conjunction with our leave-one-animal-out cross-validation procedure. We list the hyperparameters and their tuned values for each dataset in Table~\ref{tab:hyperp}. We use the same hyperparameter values as in Table~\ref{tab:hyperp} for the variant of the proposed algorithm with linear FIR filters. For the FCN and ResNet algorithms, we use the hyperparameter values prescribed in~\cite{dl4tsc}, which are shown to be almost optimal.

\begin{table} \footnotesize
\caption{The model and training hyperparameters of the proposed algorithm and their values used with each considered dataset.\\}
\label{tab:hyperp}
\centering
\begin{tabular}{|l| C{1.1cm} C{1.1cm} C{1.1cm}|}
\hline
\backslashbox{hyperparameter}{dataset} & \begin{tabular}{@{}C{1.01cm}@{}}Arm18\\6~classes\end{tabular} & \begin{tabular}{@{}C{1.01cm}@{}}Arm18\\5~classes\end{tabular} & Arm20\\
\hline
1st FIR filter length, $K_1$  & 8 & 8 & 8\\%0.783 & 
2nd FIR filter length, $K_2$  & 8 & 8 & 8\\%0.783 & 
hidden layer dimension, $L$   & 7 & 6 & 6\\
\hline
learning rate                 & 0.0005 & 0.0002 & 0.0002\\
weight decay                  & 0.004  & 0.002 & 0.002\\%0.7538 & 
batch size                    & 1024   & 1024 & 1024\\%\textbf{0.791} & 
number of training iterations & 40,000 & 60,000 & 60,000\\%0.772 & 
\hline
\end{tabular}
\end{table}

\begin{figure}
    \centering
    \subfigure[The Arm18 dataset with 6 classes.]{\includegraphics[scale=.59]{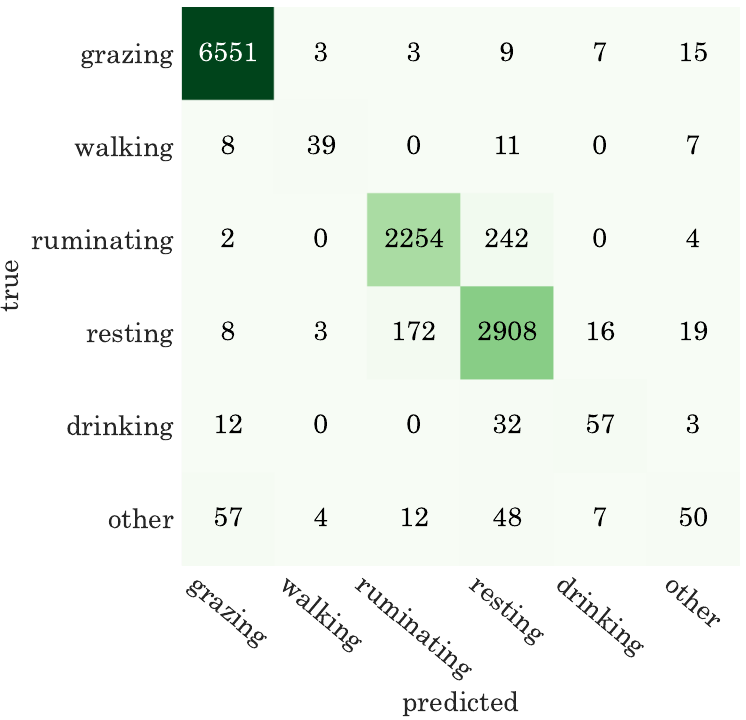}\label{cm-arm18-6c}}
    \subfigure[The Arm18 dataset with 5 classes.]{\includegraphics[scale=.59]{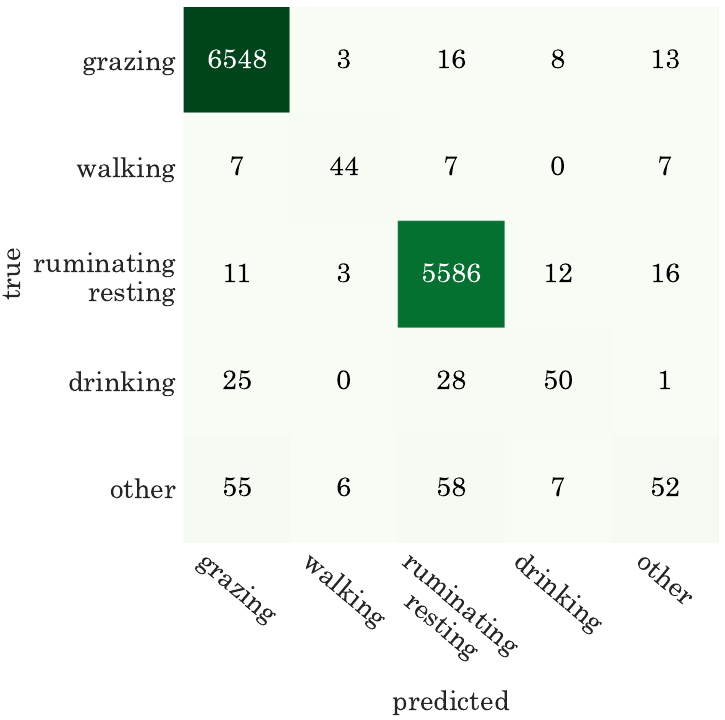}\label{cm-arm18-5c}}
    \subfigure[The Arm20 dataset.]{\includegraphics[scale=.59]{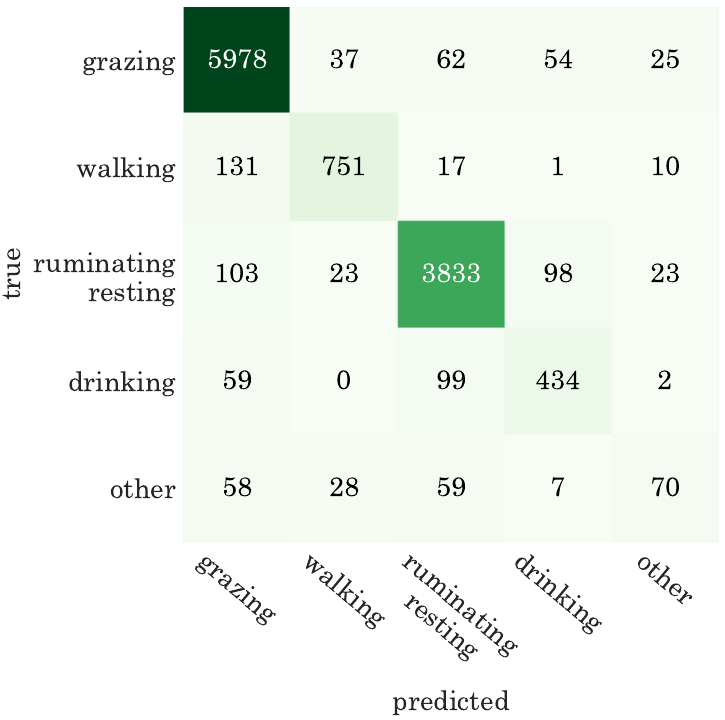}\label{cm-arm20}}
    \caption{The confusion matrices resulting from the leave-one-animal-out cross-validated evaluation of the proposed algorithm with each considered dataset.}
    \label{fig:cm}
\end{figure}

In Table~\ref{tab:mcc_cv}, we present the cross-validated MCC results for all considered algorithms and datasets. As evident in Table~\ref{tab:mcc_cv}, the proposed algorithm yields the highest MCC values compared to the other algorithms for all considered datasets. In Table~\ref{tab:mcc_cv_b}, we provide the cross-validated MCC values of the proposed algorithm for each behavior class and dataset. Fig.~\ref{fig:cm} shows the confusion matrices associated with the proposed algorithm and all considered datasets.

\begin{table} \footnotesize
\caption{The leave-one-animal-out cross-validated MCC values of the proposed algorithm and its contenders, evaluated using the considered datasets.\\}
\label{tab:mcc_cv}
\centering
\begin{tabular}{|l| C{1.1cm} C{1.1cm} C{1.1cm}|}
\hline
%& \multicolumn{3}{c|}{dataset}\\
\backslashbox{algorithm}{dataset} & \begin{tabular}{@{}C{1.1cm}@{}}Arm18\\6 classes\end{tabular} & \begin{tabular}{@{}C{1.1cm}@{}}Arm18\\5 classes\end{tabular} & Arm20\\
\hline
proposed                     & \textbf{0.9097} & \textbf{0.9568} & \textbf{0.8762}\\
proposed with linear filters & 0.9014 & 0.9467 & 0.8681\\
\cite{ca10}                  & 0.8713 & 0.9466 & 0.8662\\
FCN                          & 0.8804 & 0.9415 & 0.8713\\ 
ResNet                       & 0.9028 & 0.9478 & 0.8728\\ 
\hline
\end{tabular}
\end{table}

\begin{table} \footnotesize
\caption{The leave-one-animal-out cross-validated MCC values of the proposed algorithm for each behavior class and each considered dataset.\\}
\label{tab:mcc_cv_b}
\centering
\begin{tabular}{|l| C{1.1cm} C{1.1cm} C{1.1cm}|}
\hline
%& \multicolumn{3}{c|}{dataset}\\
\backslashbox{behavior}{dataset} & \begin{tabular}{@{}C{1.1cm}@{}}Arm18\\6~classes\end{tabular} & \begin{tabular}{@{}C{1.1cm}@{}}Arm18\\5~classes\end{tabular} & Arm20\\
\hline
grazing    & 0.9802 & 0.9780 & 0.9118\\
walking    & 0.6897 & 0.7280 & 0.8485\\
ruminating & 0.8906 & \multirow{2}{*}{0.9758} & \multirow{2}{*}{0.9099}\\
resting    & 0.8826 & & \\
drinking   & 0.5962 & 0.5556 & 0.7166\\
other      & 0.3721 & 0.4075 & 0.4038\\
\hline
overall    & 0.9097 & 0.9568 & 0.8762\\
\hline
\end{tabular}
\end{table}

\subsection{Inter-dataset Accuracy}\label{sec:ida}

Here we further assess how well a model learned using the proposed algorithm is generalizable to unseen data, i.e., data on which the model is not trained. Therefore, we evaluate the inter-dataset classification accuracy of the proposed algorithm using the five-class Arm18 and Arm20 datasets. We use the proposed algorithm to learn a behavior classification model from one dataset and evaluate it on the other dataset.

In Table~\ref{tab:mcc_cd}, we give the overall MCC values as well as those corresponding to each behavior for both cases of 1) training the proposed model on the Arm20 dataset and evaluating it on the five-class Arm18 dataset and 2) training the proposed model on the five-class Arm18 dataset and evaluating it on the Arm20 dataset. We use the same hyperparameter values as in Table~\ref{tab:hyperp}, which are in fact the same for both cases.

\begin{figure}
    \centering
    \subfigure[Training with the Arm20 dataset, test on the Arm18 dataset.]{\includegraphics[scale=.59]{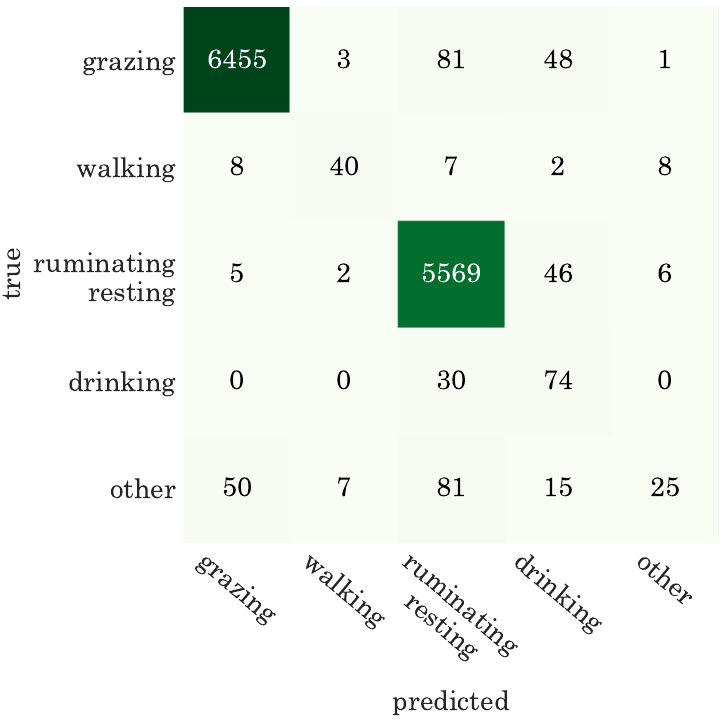}\label{arm20}}
    \subfigure[Training with the Arm18 dataset, test on the Arm20 dataset.]{\includegraphics[scale=.59]{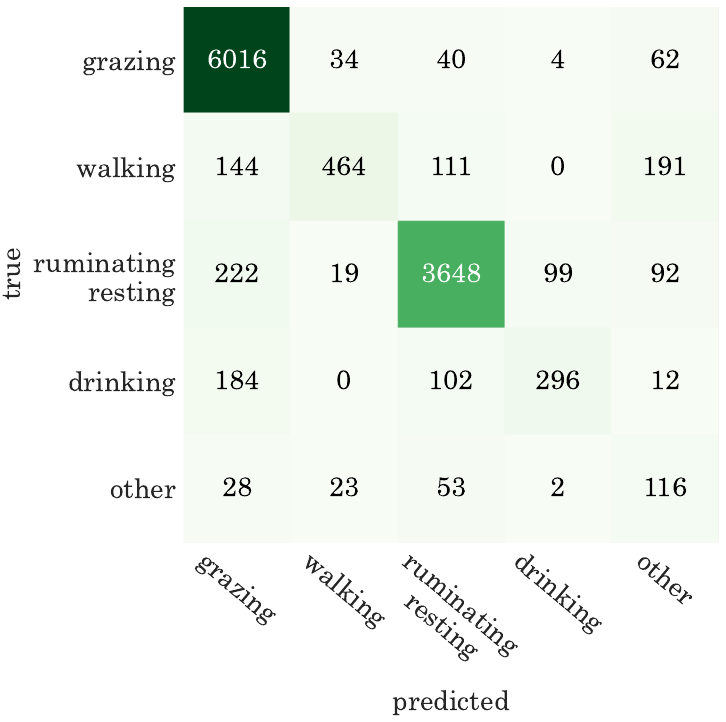}\label{arm18}}
    \caption{The confusion matrices resulting from the cross-dataset evaluation of the proposed algorithm for both considered cases.}
    \label{fig:cm_cd}
\end{figure}

Fig.~\ref{fig:cm_cd} shows the confusion matrices corresponding to the cross-dataset evaluation of the proposed algorithm using the five-class Arm18 and Arm20 datasets.

\begin{table} \footnotesize
\caption{The cross-dataset MCC values of the proposed algorithm, overall and for each behavior class, using the five-class Arm18 and Arm20 datasets.\\}
\label{tab:mcc_cd}
\centering
\begin{tabular}{|l| c c|}
\hline
\backslashbox{MCC}{training dataset\\test dataset} & \begin{tabular}{c}Arm20\\$\downarrow$\\Arm18\end{tabular} &  \begin{tabular}{c}Arm18\\$\downarrow$\\Arm20\end{tabular}\\
\hline
grazing            & 0.9688 & 0.8820\\ 
walking            & 0.6866 & 0.6423\\ 
ruminating/resting & 0.9588 & 0.8620\\ 
drinking           & 0.5285 & 0.5903\\
other              & 0.2921 & 0.3408\\ 
\hline
overall            & 0.9393 & 0.8034\\
\hline
\end{tabular}
\end{table}

Inspecting the results in Tables~\ref{tab:mcc_cv_b} and~\ref{tab:mcc_cd} shows that the models learned from both datasets generalize well to the other dataset. However, the model learned from the Arm20 dataset appears to perform better on the five-class Arm18 dataset, as opposed to the alternative. This can be due to a few factors.

First, the proportion of the less frequent classes, specifically the walking and drinking behavior classes, are significantly higher in the Arm20 dataset. Therefore, a model learned from the Arm20 dataset is expectedly more effective in classifying these behavior classes compared to a model learned from the Arm18 dataset. As seen, the classification accuracy of the walking and drinking behavior classes in the Arm20 dataset degrades considerably when a model learned from the five-class Arm18 dataset is used compared to when inter-dataset cross-validation is performed.

Second, accurate classification of the behavior classes in the Arm20 dataset appears to be more challenging compared with the five-class Arm18 dataset. This is evident from the intra-class results. Thus, classifying the Arm20 dataset using a model learned from a different dataset leads to a more noticeable loss in accuracy compared to classifying the five-class Arm18 dataset using an inter-dataset model.

\subsection{Complexity}

To perform in-situ classification of cattle behavior in real time, we implement the proposed algorithm on the embedded system of Loci using the Digital Signal Processing software library of Arm's Common Microcontroller Software Interface Standard (CMSIS). CMSIS is a vendor-independent hardware abstraction layer for microcontrollers that are based on Arm Cortex processors\footnote{\url{https://developer.arm.com/tools-and-software/embedded/cmsis}}. Particularly, we make use of the \texttt{arm\_biquad\_cascade\_df1\_f32} and \texttt{arm\_fir\_f32} functions to respectively implement the IIR and FIR filters of the proposed model.

In Table~\ref{tab:ops}, we give the number of parameters for the main parts of the proposed animal behavior classification algorithm, i.e., normalization, feature calculation, and classification. In addition, in Table~\ref{tab:ops}, we provide the number of different arithmetic/mathematical operations required for performing inference using the proposed algorithm on a single datapoint (accelerometer readings of an $N$-sample time window). The table also includes the total tally for each row when $N=256$, $K_1=K_2=8$, $F=9$, $L=7$, and $C=6$.

\begin{table*} \footnotesize
\caption{The number of parameters and the number of different operations required for performing inference on a single datapoint using the proposed animal behavior classification algorithm. The total values are for when $N=256$, $K_1=K_2=8$, $F=9$, $L=7$, and $C=6$.\\}
\label{tab:ops}
\hspace{-1.9cm}\begin{tabular}{|l| c c c|c|}
\hline
\backslashbox{complexity}{stage} & normalization & feature calculation & classification & total\\
\hline
parameters              & $6$  & $3(K_1+K_2+1)$ & $L(F+C)+C+L$ & 175\\
\hline
additions/subtractions  & $3N$ & $9N+3(N-K_1+2)K_1+3(N-K_1-K_2+2)K_2-18$ & $LF+CL$ & 14,967\\
abs. value calculations & 0    & $3(2N-K_1-K_2+2)$ & 0    & 1,494\\
multiplications         & $3N$ & $3N+3(N-K_1+1)K_1+3(N-K_1-K_2+2)K_2+6$ & $LF+CL$  & 13,431\\
tanh evaluations        & 0    & $3(N-K_1+1)$  & 0    & 747\\
ReLU operations         & 0    & 0 & $L$ & 7\\
argmax operations       & 0    & 0 & 1   & 1\\
\hline
\end{tabular}
\end{table*}

We provide the numbers related to the actual runtime complexity of performing inference on a single datapoint using the proposed algorithm in Table~\ref{tab:complx}. In this table, \enquote{text} and \enquote{rodata} refer to the ROM space occupied by the algorithm code and the model parameters, respectively. In addition, \enquote{stack} refers to the RAM space required to store all variables when running the algorithm.

As shown in Table~\ref{tab:complx}, performing inference using the proposed animal behavior classification algorithm takes $85$~milliseconds of the CPU time. This means the inference can be conveniently executed every one second. In addition, the total required memory is less than $12$KB of ROM and $10$KB of RAM while the microcontroller of Loci has access to $128$KB of flash ROM and $28$KB of RAM. Therefore, the memory requirements can be easily met.

We have verified our implementation of the proposed animal behavior classification algorithm on the embedded system of Loci using models trained on the Arm18 and Arm20 datasets during a small-scale field trial conducted with Angus beef cows in February 2022. The proposed algorithm ran smoothly in real time predicting the behavior of the cattle with a classification accuracy similar to those presented in Table~\ref{tab:mcc_cd}.

At inference time, i.e., when using the proposed algorithm to classify cattle behavior in situ, we infer the animal behavior for every window of 256 consecutive accelerometer readings (5.12s) that slides forward for 64 values (1.28s) as the new readings arrive. Therefore, the algorithm outputs the predicated behavior class every 1.28s for the last 5.12s. We count the inferred instances of each behavior class over a period of about five and half minutes (256 by 1.28s or 327.68s). We then transmit these count numbers for all behaviors to a gateway from each collar tag. This way, we avoid the costly transmission of the raw data when only the summary knowledge of animal behavior over a given time is of interest.

Each collar tag directly communicated with a gateway using a Semtech SX1272\footnote{\url{ https://www.semtech.com/products/wireless-rf/lora-core/sx1272}} long-range low-power LoRa\footnote{\url{https://lora-alliance.org/}} modem. The communication takes place at the frequency band of 916MHz with a bandwidth of 125KHz and an effective range of about 3 kilometers. In most related application scenarios, the cattle are usually within less than 3 kilometers of a gateway. The gateway is also a Loci bundled with a BeagleBone\footnote{\url{https://beagleboard.org/bone}} single-board computer that is connected to a remote server via a suitable wired or wireless link.

The payload at each round of communication that occurs every 327.68s includes six bytes for the behavior inference counts, four byte for the timestamp, and two bytes for the node ID number. Transmitting this information takes up to 1.3189s while the LoRa modem draws a current of 125mA. With a duty cycle of around 0.4\%, due to operating for 1.3189s every 327.68s, this amounts to an average current draw of about 0.5mA.

We aggregate the number of inferred instances for each behavior class over every 327.68s to optimize the efficiency of the communication. It is the longest period for which the inference counts for each behavior class can fit into a single byte. A shorter period will require more frequent communication while a longer period will entail a larger communication payload because of requiring the transmission of two or more bytes for each behavior inference count.

The CPU and all other major components of Loci including the LoRa modem draw at most 30mA on average. Therefore, the battery pack of the eGrazor collar tag with a nominal capacity of $13.4$Ah can power Loci for several weeks before needing to be recharged by the solar panels, which can provide up to $300$mA. Therefore, running the classifier and transmitting the summary knowledge of the inferred behaviors do not impose any significant burden on the available resources of Loci’s embedded system. 

\begin{table} \footnotesize
\caption{The actual memory and time complexity of the proposed animal behavior classification algorithm running on the embedded system of Loci, the eGrazor collar tag's AIoT device.\\}
\label{tab:complx}
\centering
\begin{tabular}{|c c c |c|}
\hline
%& \multicolumn{3}{c|}{dataset}\\
text & rodata & stack & CPU time\\
\hline
$10,880$ bytes & $708$ bytes & $9,550$ bytes &  $85$ ms\\
\hline
\end{tabular}
\end{table}

Note that the memory and time complexity of performing inference using the FCN and ResNet algorithms on a single datapoint is a few orders of magnitude larger compared with that of the proposed algorithm. For example, the FCN models whose MCC results are given in section~\ref{subsec:ida} have a few hundred thousand parameters taking up several megabyte of memory. In addition, a forward pass of the FCN algorithm to perform inference on a single datapoint requires around $68$ million multiplication operations. The memory and time complexity of performing inference using the ResNet algorithm is more than double that of the FCN algorithm.

\section{Interpretation}\label{sec:intp}

We provide some insights into the proposed animal behavior classification algorithm, particularly, the features that it extracts from the triaxial accelerometer readings in an end-to-end manner by analyzing the statistical and spectral properties of the data and their relationships with the features. We choose the Arm20 dataset for this purpose as it is less unbalanced compared with the Arm18 dataset in terms of the prevalence of different behavior classes.

\subsection{Features}

\begin{figure}
    \centering
    \subfigure[The $x$ axis.]{\includegraphics[scale=.545]{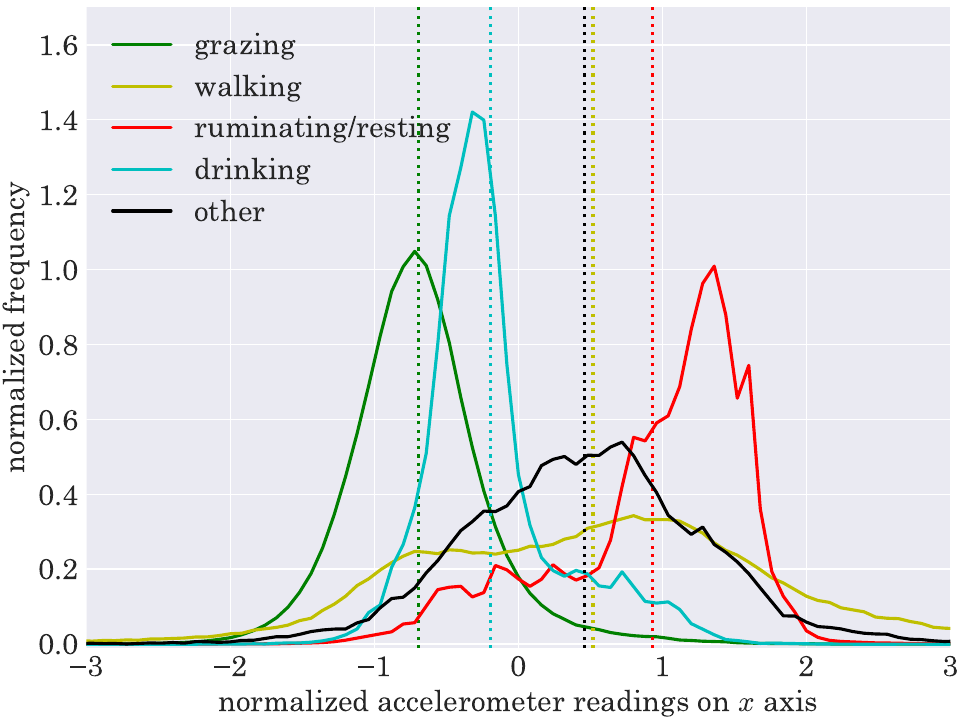}\label{histax}}
    \subfigure[The $y$ axis.]{\includegraphics[scale=.545]{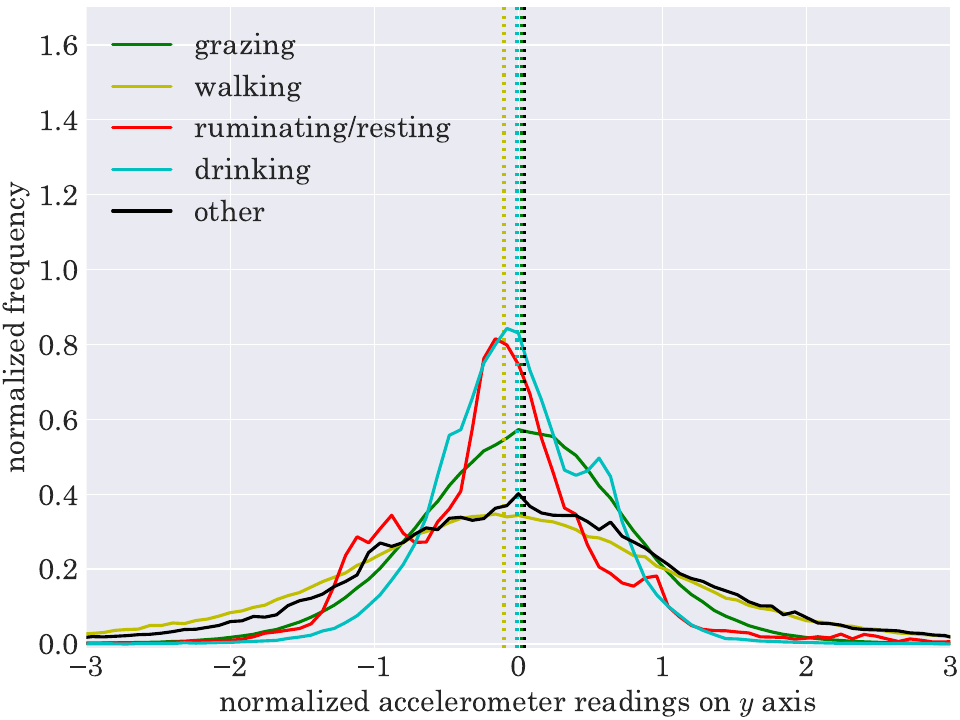}\label{histay}}
    \subfigure[The $z$ axis.]{\includegraphics[scale=.545]{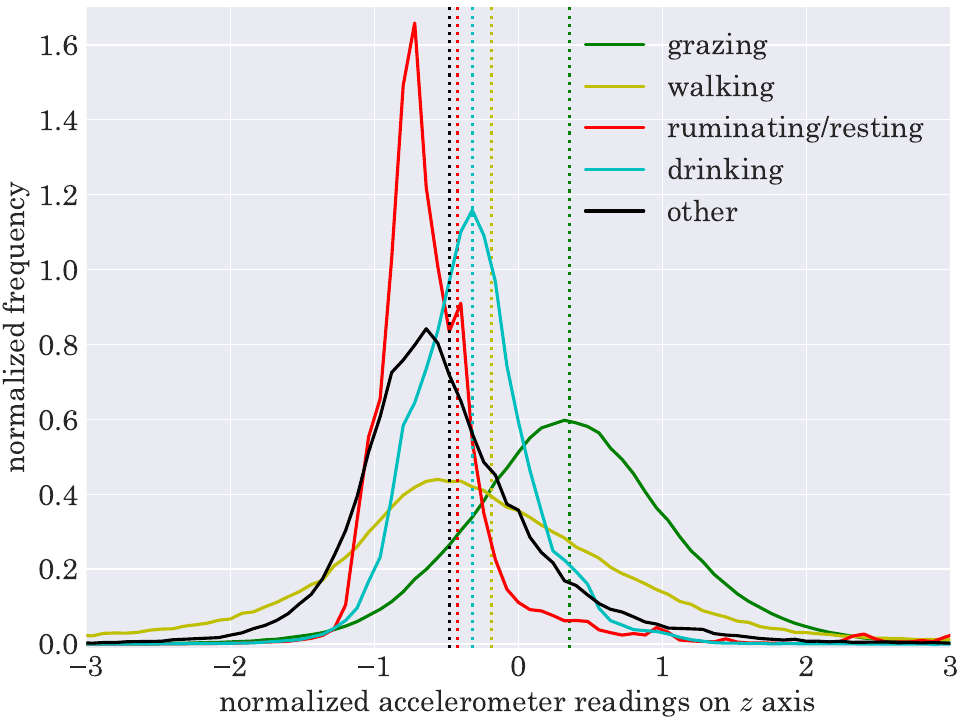}\label{histaz}}
    \caption{\small{The histograms of the normalized accelerometer readings of the Arm20 dataset for each class and spatial axis.}}
   \label{hists}
\end{figure}

In Fig.~\ref{hists}, we plot the histograms of the normalized accelerometer readings, i.e., $\bar{\mathbf{a}}_d$, $d\in\{x,y,z\}$, for each behavior class and spatial axis. Each dashed vertical line in Fig.~\ref{hists} indicates the mean value of its corresponding behavior class with the same color.

We observe in Fig.~\ref{hists} that the means corresponding to different behavior classes, especially those for the $x$ axis, are rather distinct. Therefore, they can be useful for discriminating the behavior classes. The mean values are directly related to the orientation of the tag and hence the head pose of the animal wearing the tag. Considering the behaviors of interest, the head pose can carry significant information in regards to the animal's behavior. The mean features, i.e., $f_{1d}$, $d\in\{x,y,z\}$, are meant to capture this information.

\begin{figure}
    \centering
    \subfigure[The $x$ axis.]{\includegraphics[scale=.545]{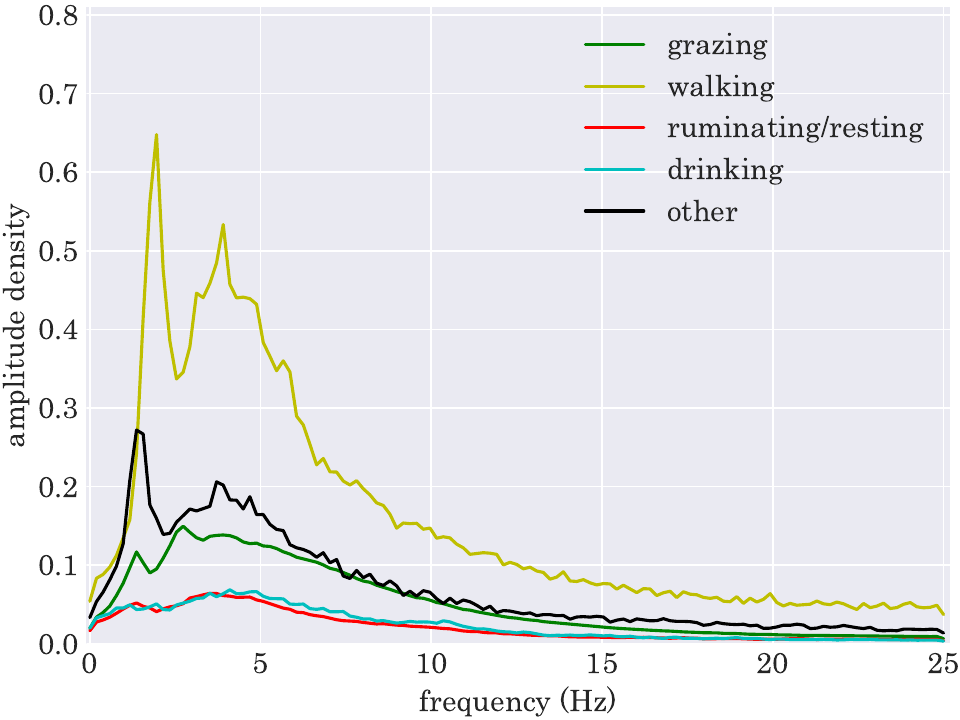}\label{asdx}}
    \subfigure[The $y$ axis.]{\includegraphics[scale=.545]{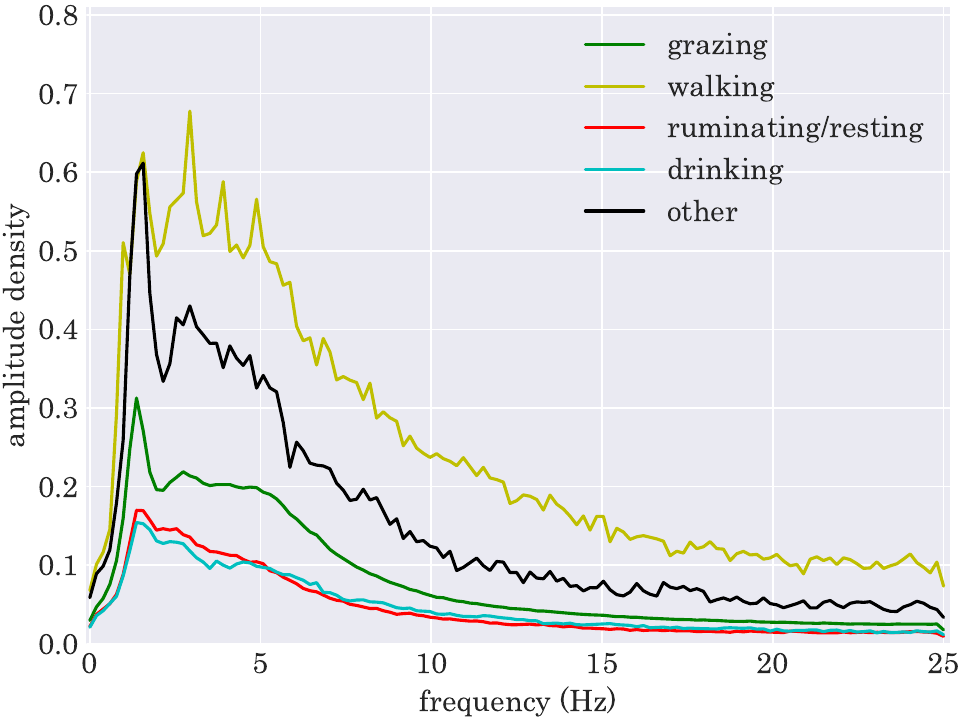}\label{asdy}}
    \subfigure[The $z$ axis.]{\includegraphics[scale=.545]{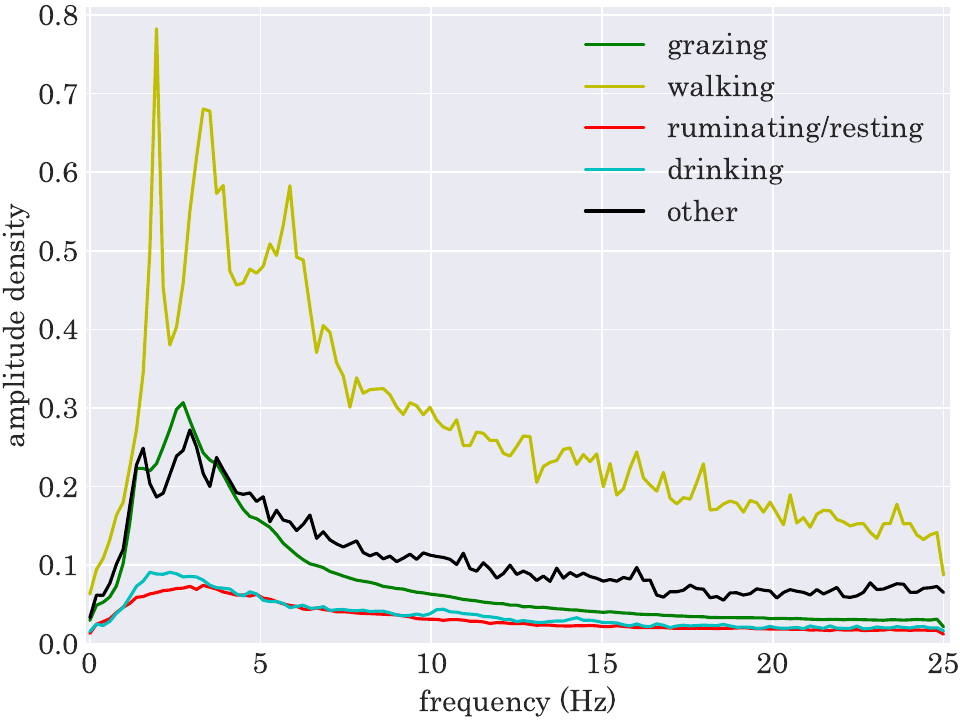}\label{asdz}}
    \caption{\small{The amplitude density functions of the normalized and IIR-filtered accelerometer readings for all behavior classes and spatial axes, averaged over all datapoints of the Arm20 dataset.}}
   \label{asd}
\end{figure}

In Fig.~\ref{asd}, we plot the amplitude spectral density (ASD) functions of the normalized and IIR-filtered accelerometer readings, i.e., $\tilde{\mathbf{a}}_d$, $d\in\{x,y,z\}$, for all behavior classes and spatial axes. The ASDs are averaged over all datapoints ($N$-sample segments) of the Arm20 dataset. The ASD function is the square-root of the power spectral density function. It represents how the power of the accelerometer readings within the $N$-sample segments are on-average distributed over the spectral range of zero to $25$Hz (the Nyquist frequency that is half of the sampling frequency) for each behavior class and axis.

Fig.~\ref{asd} shows that the overall power of the IIR-filtered accelerometer readings (with the effect of gravity/head pose removed) can be a good distinguishing factor for most behavior classes. We use the features $f_{2d}$, $d\in\{x,y,z\}$, to capture this information that relates to the intensity of animal's body movements. We use the mean-absolute value instead of the standard deviation to quantify the power because of its superior numerical properties such as being less computationally demanding and more robust to noise and outliers.

\begin{figure}
    \centering
    \subfigure[The $x$ axis.]{\includegraphics[scale=.545]{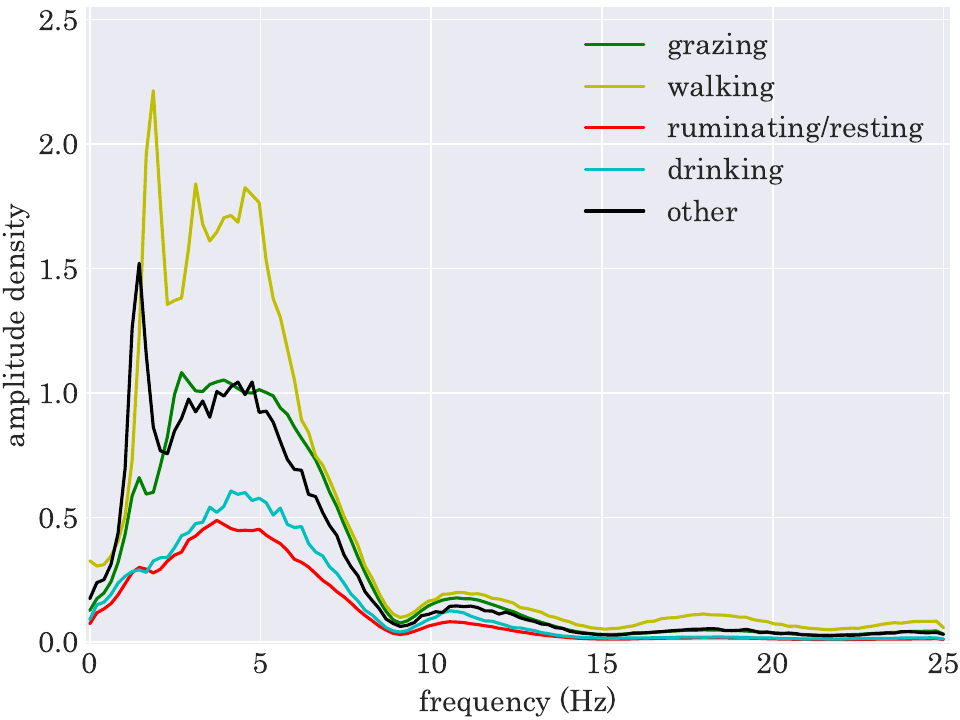}\label{asdfx}}
    \subfigure[The $y$ axis.]{\includegraphics[scale=.545]{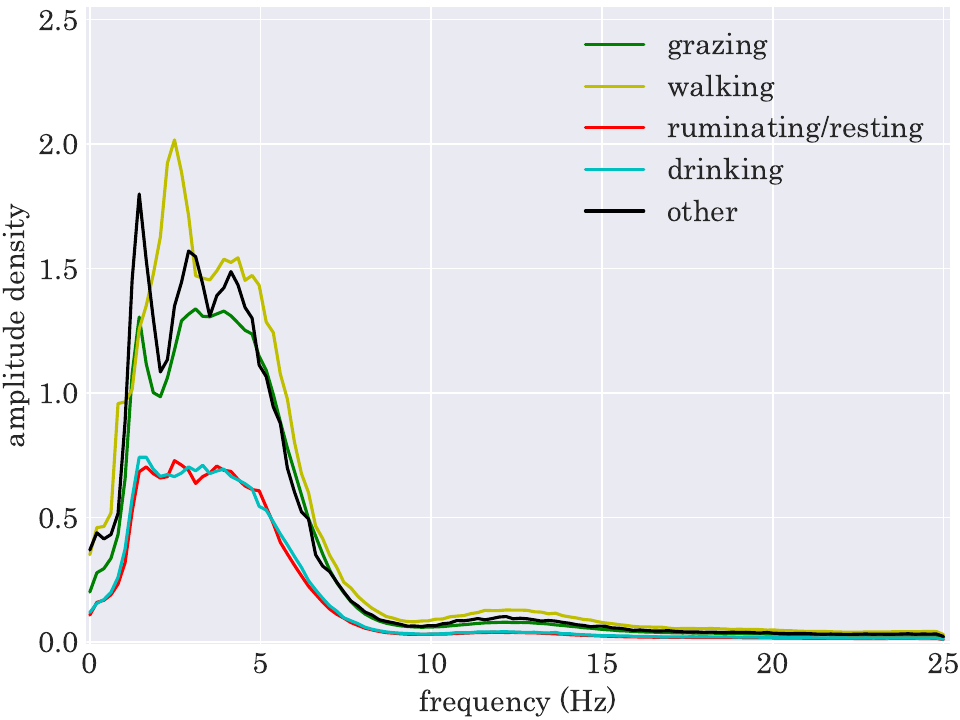}\label{asdfy}}
    \subfigure[The $z$ axis.]{\includegraphics[scale=.545]{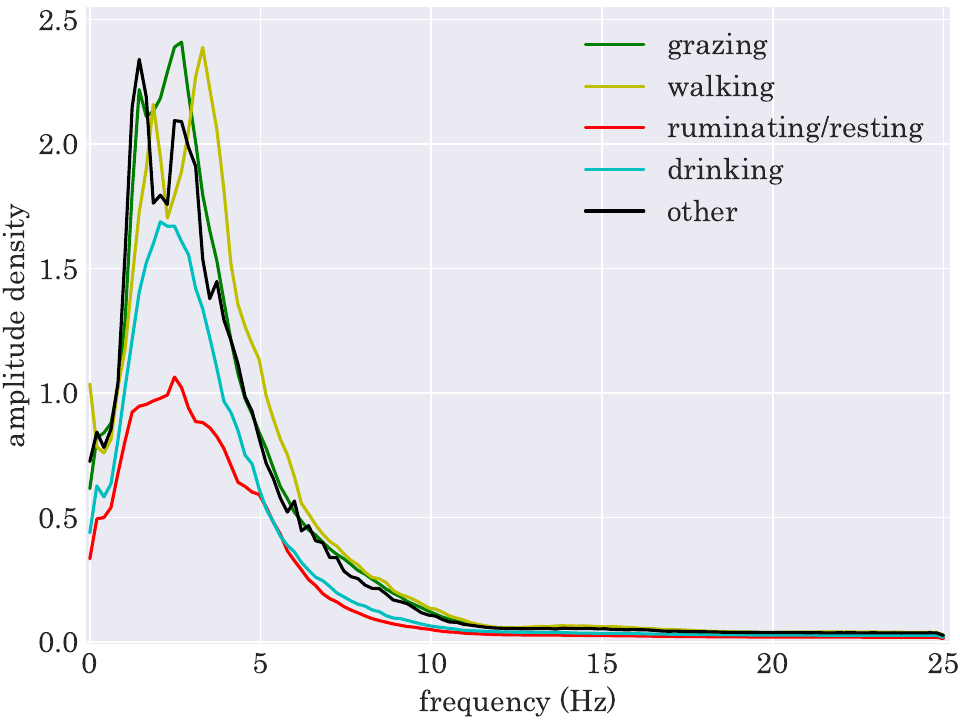}\label{asdfz}}
    \caption{\small{The amplitude density functions of the normalized and filtered accelerometer readings for all classes and axes, averaged over all datapoints of the Arm20 dataset.}}
   \label{asdf}
\end{figure}

In Fig.~\ref{asdf}, we plot the ASD of the nonlinear-filtered accelerometer readings that are used to calculate $f_{3d}$, $d\in\{x,y,z\}$, i.e.,
\begin{equation}
\check{\textbf{a}}_d=\tanh\left(\tilde{\mathbf{a}}_d*\mathbf{h}_{1d}\right)*\mathbf{h}_{2d},\ d\in\{x,y,z\},
\end{equation}
for all behavior classes and spatial axes when the model parameters are learned for the Arm20 dataset.

We make two major observations from Fig.~\ref{asdf}. First, the ruminating/resting and drinking behavior classes have similar overall powers in the $z$ axis for pre-nonlinear-filtered accelerometer readings as seen in Fig.~\ref{asdz}, specifically in comparison with the other classes. However, after the nonlinear filtering, as shown in Fig.~\ref{asdfz}, the filtered values associated with the two behavior classes have substantially different total powers. This means $f_{3z}$ can help distinguish the drinking behavior from the ruminating/resting behavior and consequently from the other behaviors. Note that drinking is a relatively rare behavior and generally hard to classify accurately. Second, the high-frequency spectral components of the accelerometer readings, i.e., over $10$Hz, appear to be mostly suppressed by the nonlinear filters trained on the Arm20 dataset to classify animal behavior. This is justifiably beneficial as the considered cattle behaviors are expected to have acceleration signatures that predominately fall in the frequency range lower than $10$Hz. The higher-frequency components are most likely due to observational noise/error.

\begin{figure}
    \centering
    \subfigure[The frequency responses of the first set of FIR filters, $\mathbf{h}_{1d}$, $d\in\{x,y,z\}$.]{\includegraphics[scale=.545]{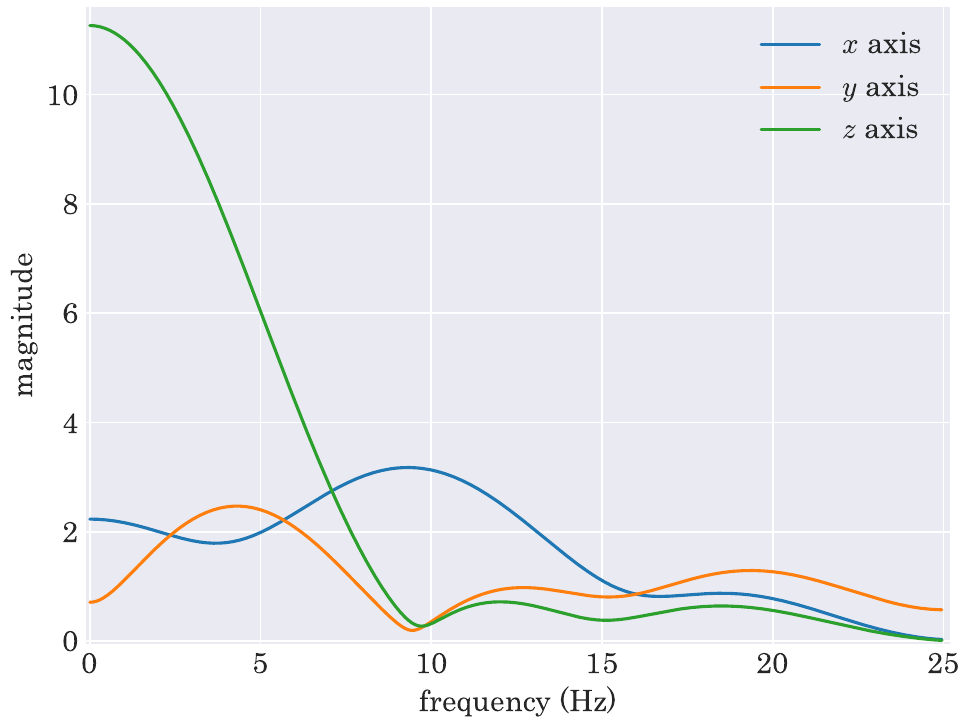}\label{frqz1}}
    \subfigure[The frequency responses of the second set of FIR filters, $\mathbf{h}_{2d}$, $d\in\{x,y,z\}$.]{\includegraphics[scale=.545]{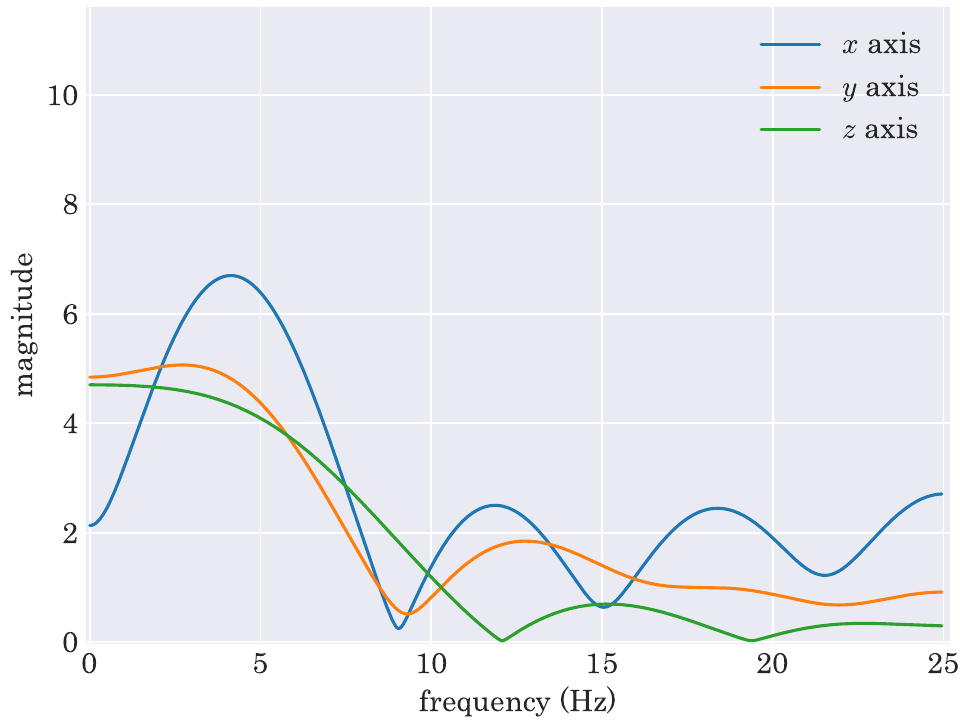}\label{freqz2}}
    \caption{\small{The frequency responses of the FIR filters associated with all spatial axes in the proposed algorithm, learned from the Arm20 dataset.}}
   \label{freqz}
\end{figure}

In Fig.~\ref{freqz}, we plot the frequency responses of the learned FIR filters, i.e., $\mathbf{h}_{1d}$ and $\mathbf{h}_{2d}$, $d\in\{x,y,z\}$. The inclusion of these plots is only for the sake of illustration as the FIR filters in the proposed algorithm form a set of nonlinear filters together with the utilized element-wise $\mathrm{tanh}$ activation function. Frequency response is undefined for these nonlinear filters, which result in the filtered values with the ASD functions shown in Fig.~\ref{asdf}.

\subsection{Feature space}

To gain more insights into the inner-workings of the proposed algorithm, we visualize the feature space associated with the Arm20 dataset in two embedding dimensions using the $t$-distributed stochastic neighbor embedding (tSNE) algorithm~\cite{tSNE}. To this end, we calculate the features, i.e., $f_{i,d}$, $i\in\{1,2,3\}$ $\&$ $d\in\{x,y,z\}$, for the entire Arm20 dataset using the parameters of a model trained on the same dataset to classify its behavior classes. The tSNE algorithm preserves the local structure of the subspace constituted by the features while projecting it onto a lower-dimensional space. It does not necessarily preserve the global structure of the data.

\begin{figure}
    \centering
    \subfigure[Using $f_{1d}$, $f_{2d}$, $f_{3d}$, $d\in\{x,y,z\}$.]{\includegraphics[scale=.545]{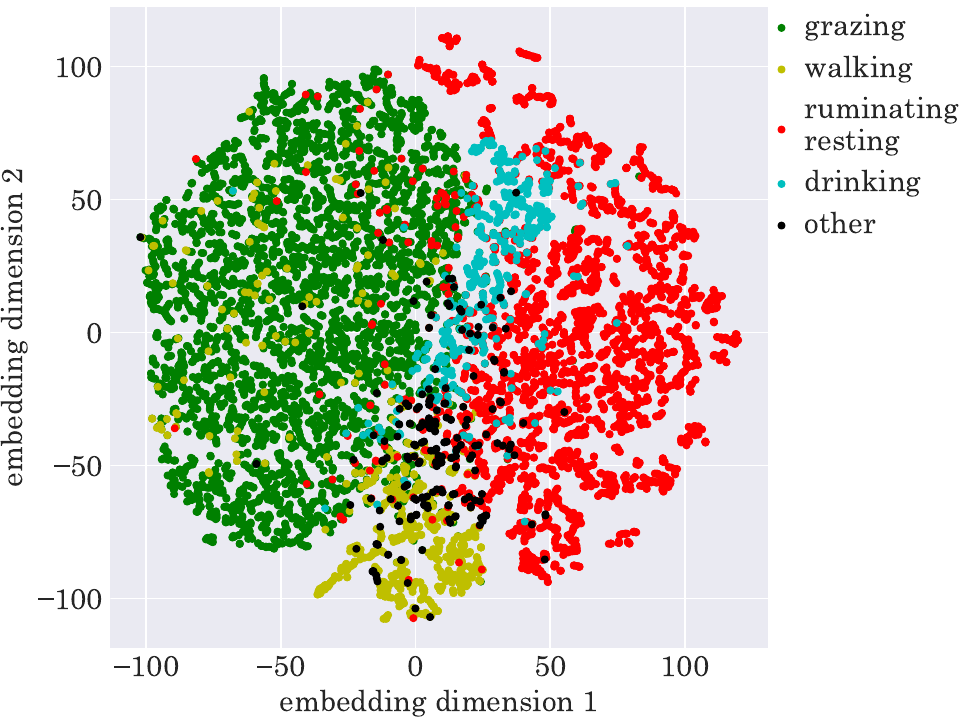}\label{tsne-9}}
    \subfigure[Using $f_{1d}$, $f_{2d}$, $d\in\{x,y,z\}$.]{\includegraphics[scale=.545]{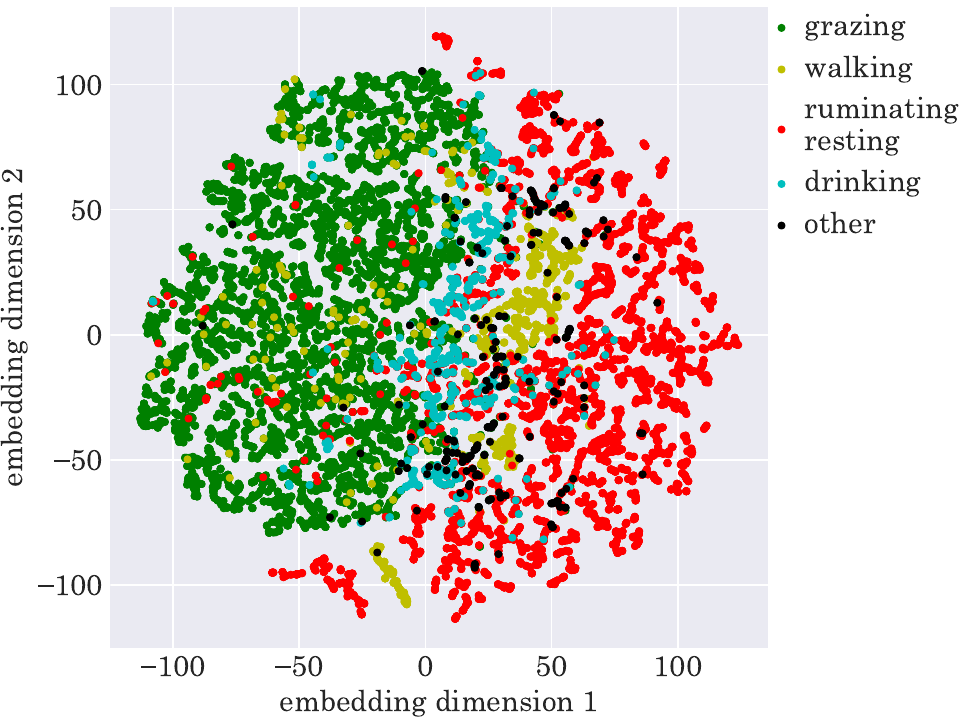}\label{tsne-6}}
    \subfigure[Using nine features calculated through the model trained on the Arm18 dataset.]{\includegraphics[scale=.545]{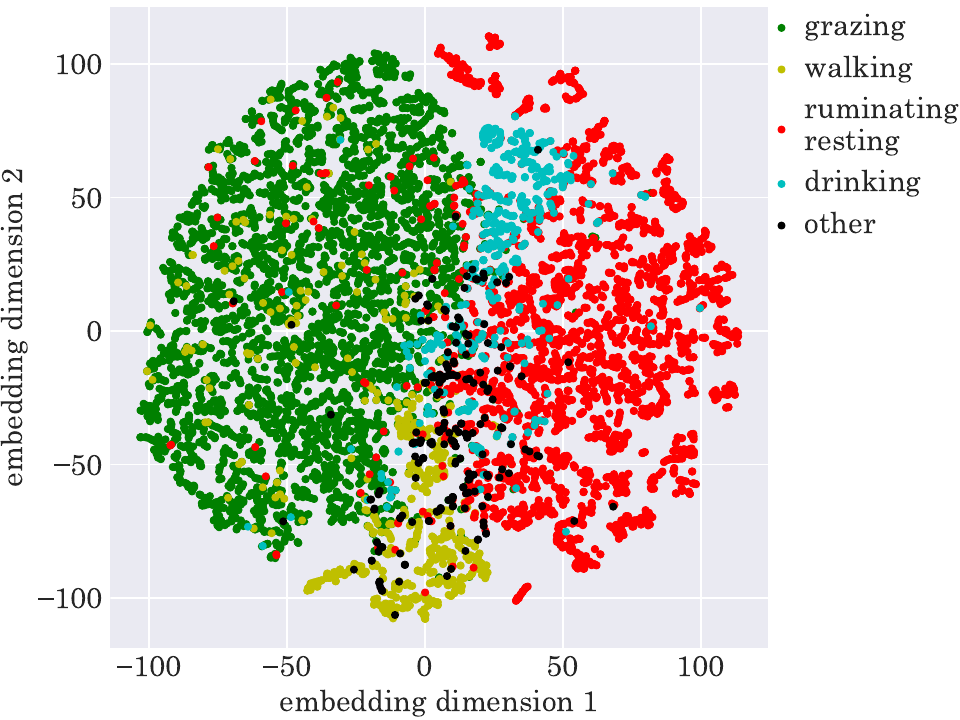}\label{tsne-o}}
    \caption{\small{Visualization of the Arm20 dataset in the feature space using the tSNE algorithm.}}
    \label{tsne}
\end{figure}

Fig.~\ref{tsne-9} is a visualization of the feature space of the Arm20 dataset using all nine features while Fig.~\ref{tsne-6} is another visualization using only the first six features, i.e., $f_{1d}$, $f_{2d}$, $d\in\{x,y,z\}$. Each dot in Fig.~\ref{tsne} represents a datapoint and is colored according to its corresponding behavior class. It is clear from Figs.~\ref{tsne-9} and~\ref{tsne-6} that the additional three features, i.e., $f_{3d}$, $d\in\{x,y,z\}$, help datapoints belonging to the same class cluster around each other better hence facilitate the classification and improve accuracy. This is more prominent for the less frequent behavior classes, i.e., walking, drinking, and other.

In Fig.~\ref{tsne-o}, we visualize the feature space of the Arm20 dataset when the nine features are calculated using a model trained on the Arm18 dataset. The clusters corresponding to different behavior classes are similarly distinguishable in Figs.~\ref{tsne-9} and~\ref{tsne-o}. This can partially explain the favorable inter-dataset generalizability of the proposed model observed in section~\ref{sec:ida}.

\section{Discussion}

The work presented in this paper is a continuation of our previous work in~\cite{ca10} that improves it in several aspects. First and foremost, our approach here is based on end-to-end learning where the feature extractor parameters are optimized jointly with the classifier parameters. In~\cite{ca10}, feature extraction and classification are done separately. Moreover, in this work, we use three new features that help enhance performance, particularly by facilitating the classification of less frequent behaviors such as drinking. Unlike in~\cite{ca10}, here, we also normalize the accelerometer readings before calculating the features using them. This improves the classification accuracy as well as the training speed.

We use a single set of IIR and nonlinear filters to calculate the features in the proposed algorithm. However, it is straightforward to extend the proposed algorithm to calculate more features using multiple sets of filters. In our experiments with the considered datasets, we did not find any significant improvement in classification accuracy when using more filters. We did not observe any benefit in using nonlinear filters that have more than two FIR filters in tandem either. Nor did we witness any benefit in having more than one hidden layer in the MLP classifier.

In the proposed algorithm, we treat the accelerometer readings of three spatial axes independently; hence, the FIR filters are akin to depthwise convolutions with no bias. We have considered using two-dimensional convolutions or adding pointwise convolutions to take into account possible inter-channel information. However, despite the significant increase in complexity, there was no gain in classification accuracy. Addition of bias to the FIR filters was not beneficial either.

We have considered using batch normalization, dropout regularization, and skip connections in the model underpinning the proposed algorithm or its training. However, none led to any improvement in the classification accuracy.

The $\mathrm{tanh}$ activation function used within the nonlinear filters of the proposed algorithm results in substantially higher classification accuracy compared with using ReLU or sigmoid (logistic) activation functions. However, its implementation on embedded systems is resource intensive. In future work, we will consider replacing it with a less complex approximation or implementing it more efficiently without incurring any significant loss of accuracy.

The proposed algorithm does not show any sign of overfitting to the considered datasets when using the hyperparameter values given in Table~\ref{tab:hyperp}. On the other hand, the FCN and ResNet algorithms overfit in every scenario regardless of the choice of the hyperparameter values as they are large enough to memorize the uninformative and irrelevant patterns in the training data that are likely due to noise or nuisance factors. Therefore, when training these CNN-based model, we treat the number of training iterations as a hyperparameter and tune it through cross-validation. We do not need to limit the number of the training iterations of the proposed model to prevent it from overfitting the training set in our experiments with the considered datasets. The iteration numbers in Table~\ref{tab:hyperp} indicate when the convergence occurs and further training does not reduce the aggregate cross-entropy loss.

Modularity and flexibility of the modern deep neural networks, enabled by their layered structure that can incorporate nonlinear functions and transformations, have led to their widespread successful use in several applications that demand learning approximations to complex nonlinear mapping functions. However, the advantages of the deep neural networks come at the expense of high nonlinearity and nonconvexity of the associated optimization objective functions. This has made it practically impossible to analyze the performance of deep learning models theoretically or predict their accuracy from an analytical point of view~\cite{DLb}. Interpreting deep learning models and explaining their performance are areas of active research~\cite{DLi}.

In section~\ref{sec:intp}, we attempt to interpret the underlying deep neural networks architecture of the proposed algorithm and explain how works. Explaining alternate architectures that are outperformed by the proposed algorithm and why that is the case is hard if possible at all. Therefore, we only present and examine the performance of the proposed algorithm and suffice with mentioning some notable alternatives, which we have investigated, in the above paragraphs. Architectural hyperparameters, such as the number of layers, the number of filters in each layer, and the type of activation functions, are generally determined through cross-validation and limited, often greedy, search in the space of feasible hyperparameter values. Finding the optimal values for the hyperparameters is impractical as it requires combinatorial optimization with typically prohibitive time and space complexity.

Learning animal behavior classification models that perform well on rare behavior classes such as drinking is intrinsically challenging. This is mainly because the amount of training data available for such behaviors is limited. The annotation of these behaviors is also hard as they may be overlooked or mistaken due to occurring sporadically and in short intervals. Some grazing cattle may not drink water from any water trough for several days depending on the circumstances. When a classifier is learned from data that does not represent the entire subspace corresponding to some classes, its accuracy and confidence will inevitably be affected adversely, particularly, with respect to the inadequately characterized classes.

Paucity of data for uncommon behavior classes makes the training dataset highly imbalanced. This is certainly unideal. However, in our experience, the class imbalance is not the main culprit for inferior accuracy of classifying the rare behaviors. Rather, the scarcity of training data for these behavior classes is to be blamed. We have explored using various methods for balancing our datasets, such as undersampling, oversampling, weighting the datapoints by the inverse of the frequency of their associated classes, and synthesizing new datapoints using the synthetic minority oversampling technique~\cite{smote}. However, we have not observed any meaningful improvement in classification accuracy using models learned from the resultant artificially balanced datasets. Learning accurate classification models for rare behaviors is a subject of our ongoing research.

In our current implementation of the proposed algorithm on the embedded system of Loci, we use $32$-bit floating-point parameters and variables and performs all the required mathematical operations with these numbers using the corresponding floating-point operations. In future work, we will consider using quantization to reduce the number of floating-point operations and consequently accelerate the in-situ inference procedure.

In our annotated datasets, each datapoint belongs to only one behavior class for the entirety of its temporal dimension. On the other hand, at inference time, during every consecutive 256 accelerometry readings or 5.12 seconds, the animal may not necessarily exhibit a single behavior as it inevitably switches between behaviors at arbitrary occasions. The ratio of time segments over which inference is performed while more than one behavior occur can be decreased using a smaller segment size. However, in practice, such instances cannot be eliminated for being a fundamental limitation of performing inference on segments of any time-series data. One potential way to tackle this limitation is to use running statistics instead of time-windowed statistics. Another possible way is through running an online time-series change detection algorithm alongside the behavior classification algorithm. We will study these possible alternative solutions and the associated challenges and opportunities in our future research.

\section{Conclusion}

We developed a new algorithm for animal behavior classification using triaxial accelerometry data. The proposed model can be trained in an end-to-end manner and implemented on the embedded system of our purpose-built AIoT device to perform animal behavior classification in situ and in real time. The proposed algorithm computes three sets of features that capture information from triaxial accelerometry data regarding the animal behavior in insightful ways. It uses an MLP to classify the calculated features. When evaluated using two datasets collected via real-world animal trials, the proposed algorithm delivers classification accuracy that is superior to that of two state-of-the-art CNN-based classifiers while it incurs substantially lower memory and time complexity.

\section*{Acknowledgment}

We would like to thank the following technical staff who were involved in the research at CSIRO FD McMaster Laboratory Chiswick: Alistair Donaldson and Reg Woodgate with NSW Department of Primary Industries, and Jody McNally and Troy Kalinowski with CSIRO Agriculture and Food.

\end{document}